 \documentclass[final,5p,times,twocolumn]{elsarticle}
 \usepackage[table]{xcolor}
\usepackage[utf8]{inputenc}
\usepackage{mathtools}
\usepackage{amsmath,amssymb,amsfonts}
\usepackage{algorithmic}
\usepackage{graphicx}
\usepackage{textcomp}
\usepackage{xcolor}
\usepackage{ulem}
\usepackage{algorithmic,comment}
\usepackage[ruled,vlined]{algorithm2e}
\usepackage[T1]{fontenc}
\usepackage[utf8]{inputenc}
\usepackage{verbatim}
\usepackage{soul} 
\usepackage[hyphenbreaks]{breakurl}
 \usepackage{float}
\usepackage{hyperref}
\usepackage[hyphenbreaks]{breakurl}
\usepackage{multirow}
\usepackage{adjustbox}

\newcommand{\blt}[0]{\color{black}}

\newcommand{\ar}[1]{\textcolor{black}{#1}}
\newcommand{\xc}[1]{\textcolor{green}{#1}}

\newcommand{\arr}[1]{\textcolor{red}{\textit{Razi: #1}}}

\usepackage{enumerate}
\usepackage{colortbl}
\usepackage{booktabs}
\usepackage{array}
\usepackage{setspace}

\usepackage[switch]{lineno}

\def \includeAppendix{1}
\def \incYes{1}

\makeatletter
\def\UrlAlphabet{%
      \do\a\do\b\do\c\do\d\do\e\do\f\do\g\do\h\do\i\do\j%
      \do\k\do\l\do\m\do\n\do\o\do\p\do\q\do\r\do\s\do\t%
      \do\u\do\v\do\w\do\x\do\y\do\z\do\A\do\B\do\C\do\D%
      \do\E\do\F\do\G\do\H\do\I\do\J\do\K\do\L\do\M\do\N%
      \do\O\do\P\do\Q\do\R\do\S\do\T\do\U\do\V\do\W\do\X%
      \do\Y\do\Z}
\def\UrlDigits{\do\1\do\2\do\3\do\4\do\5\do\6\do\7\do\8\do\9\do\0}
\g@addto@macro{\UrlBreaks}{\UrlOrds}
\g@addto@macro{\UrlBreaks}{\UrlAlphabet}
\g@addto@macro{\UrlBreaks}{\UrlDigits}
\makeatother

\usepackage{amssymb}

\begin{document}

\begin{frontmatter}



\title{Network-level Safety Metrics for Overall Traffic Safety Assessment: A Case Study\tnoteref{t1}}
\tnotetext[t1]{This material is based upon the work supported by the National Science Foundation under
Grant No. 2008784 and the Arizona Commerce Authority under Institute of Automated Mobility (IAM) project.}

\author[inst1]{Xiwen Chen}
\author[inst1]{Hao Wang}
\author[inst1]{Abolfazl Razi\corref{cor1}}\ead{arazi@clemson.edu}
\author[inst2]{Brendan Russo}
\author[inst3]{Jason Pacheco}
\author[inst4]{John Roberts}
\ifx \includeAppendix \incYes
\author[inst5]{Jeffrey Wishart}
\fi
\author[inst6]{Larry Head}
\author[inst3]{Alonso Granados Baca}
\address[inst1]{School of Computing, Clemson University, Clemson, SC 29631}
\address[inst2]{Department of Civil Engineering, Northern Arizona University, Flagstaff, AZ 86011}
\address[inst3]{Department of Computer Science, The University of Arizona, Tucson, AZ 85721}
\address[inst4]{Arizona Department of Transportation, Phoenix, AZ 85009}
\ifx \includeAppendix \incYes
\address[inst5]{Exponent Co, Phoenix, AZ 85721} 
\fi
\address[inst6]{Systems and Industrial Engineering, The University of Arizona, Tucson, AZ 85027}

\cortext[cor1]{Corresponding author}


\begin{abstract}

\ar{Driving safety analysis has recently experienced unprecedented improvements thanks to technological advances in precise positioning sensors, artificial intelligence (AI)-based safety features, autonomous driving systems, connected vehicles, high-throughput computing, and edge computing servers. Particularly, deep learning (DL) methods empowered volume video processing to extract safety-related features from massive videos captured by roadside units (RSU).
Safety metrics are commonly used measures to investigate crashes and near-conflict events. However, these metrics provide limited insight into the overall network-level traffic management. On the other hand, some safety assessment efforts are devoted to processing crash reports and identifying spatial and temporal patterns of crashes that correlate with road geometry, traffic volume, and weather conditions. This approach relies merely on crash reports and ignores the rich information of traffic videos that can help identify the role of safety violations in crashes.} 

\ar{To bridge these two perspectives, we define a new set of network-level safety metrics (NSM) to assess the overall safety profile of traffic flow by processing imagery taken by RSU cameras. Our analysis suggests that NSMs show significant statistical associations with crash rates. This approach is different than simply generalizing the results of individual crash analyses, since all vehicles contribute to calculating NSMs, not only the ones involved in crash incidents. This perspective considers the traffic flow as a complex dynamic system where actions of some nodes can propagate through the network and influence the crash risk for other nodes.    
The analysis is carried out using six video cameras in the state of Arizona along with a 5-year crash report obtained from the Arizona Department of Transportation (ADOT). The results confirm that NSMs modulate the baseline crash probability. Therefore, online monitoring of NSMs can be used by traffic management teams and AI-based traffic monitoring systems for risk analysis and traffic control.We also provide a comprehensive review of \textit{surrogate safety metrics} (SSM) in the \ref{sec:safety-metrics}.  } 



\end{abstract}

\begin{keyword}
Deep Learning \sep Driving Safety Analysis \sep Safety Metrics \sep  Autonomous Vehicles.

\end{keyword}

\end{frontmatter}


\section{Introduction}\label{sec:intro}

Vehicular technology has witnessed key milestones in recent years. 
Most cars are heavily equipped with advanced visual and radio sensors, cameras, control units, and artificial-intelligence (AI)-platforms that make driving safer and more convenient than ever. Electric vehicles (EVs) equipped with automated driving systems (ADS) have achieved higher levels of autonomy and continue to expand their territory in the global car market\cite{iea_2021}. Crowd-souring and connected vehicle (CV) services have been utilized to improve the overall operation of the vehicular networks through data and model sharing. For instance, Uber Advanced Technologies Group has recently proposed a unified deep learning (DL) framework that assists automated vehicles (AVs) to map, perceive, predict, and plan sequentially \ar{to enhance driving} safety\cite{casas2021mp3}.  
Another example is Tesla which employs a cluster with 5,760 A100 GPUs to conveniently train their multi-modality neural network with a 1.5 petabytes dataset\cite{workshop_on_autonomous_driving}. 

\ar{Intel's mobile eye \cite{mobileeye,mobileye2020adas}, Google's Waymo one \cite{waymo,xu2021spg}, and Nvidia's Drive \cite{omniverse2022,nvidia2022selfdrive} are other examples of using DL-based AI platforms for autonomous and safe driving application.}


The use of AI platforms is not limited to car manufacturing. It indeed made revolutionary changes to traffic monitoring and control systems, and roadside infrastructures. Particularly, web-based high-performance computing (HPC), and vehicular edge computing (VEC) servers with graphics/tensor processing units (GPU/TPUs) have made volume data aggregation and processing, more feasible than ever \cite{liu2021vehicular}. 

Despite these technological advances, driving safety still remains one of the key challenges of today's society. Statistics show that the mortality of motor-vehicle related injuries has been almost constant 
in years 2015 to 2019 in the US \cite{iihs_2019}, while the car crash fatalities even have been increased since the COVID-19 pandemic started \cite{media_2021,nhtsa_2021}. 
This is a global issue, and about 1.3 million people die by car accidents worldwide, and millions are injured every year according to the world health organization (WHO) \cite{world_health_organization}. 
These statistics reveal that modern technology has not yet been fully utilized to prevent avoidable accident casualties and fatalities. 

\begin{figure}[]
\begin{center}
\centerline{\includegraphics[width=0.8\columnwidth]{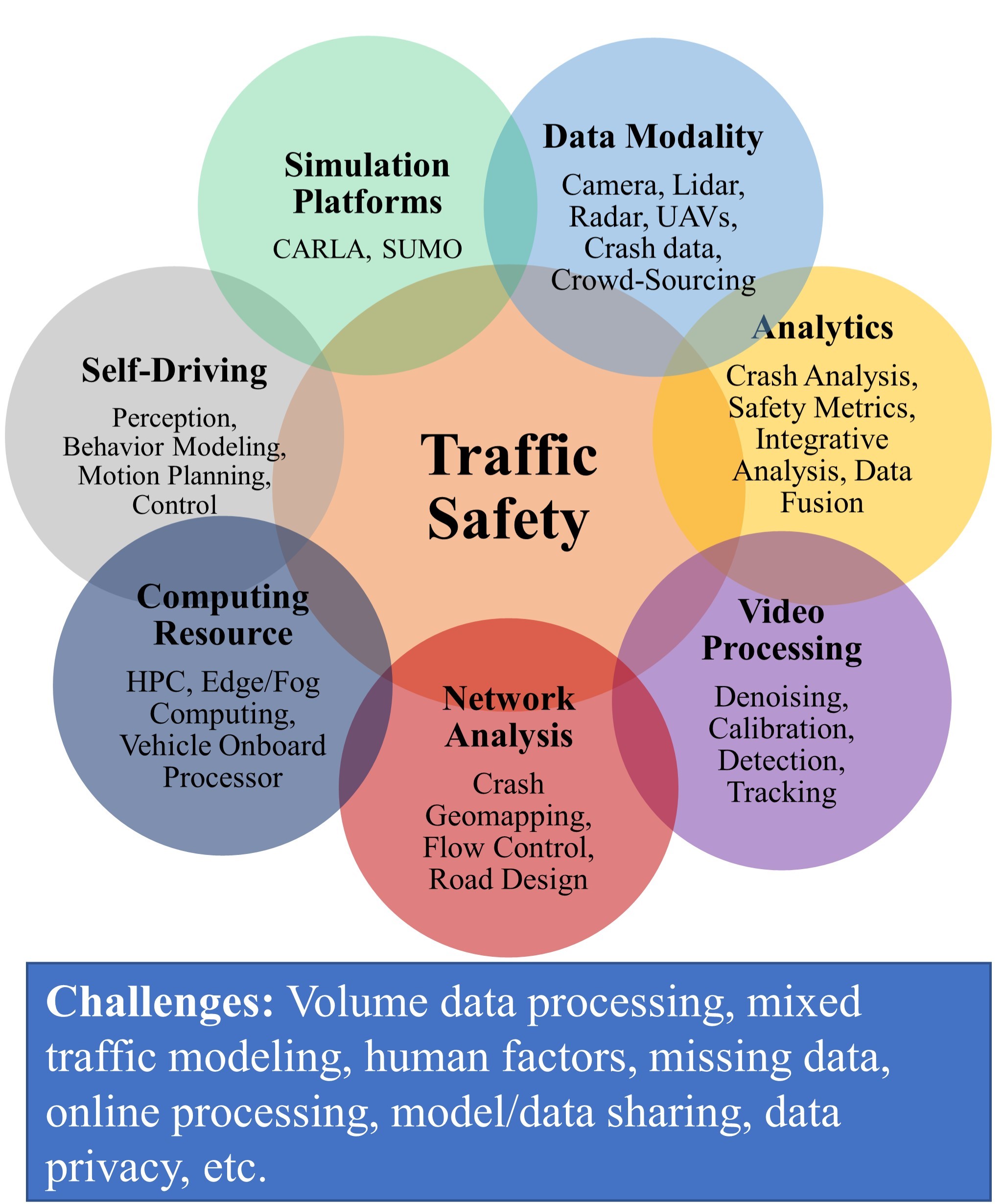}}
\caption{Different aspects, tools, opportunities, and challenges related to traffic safety.}
\label{fig:topics}
\end{center}
\end{figure}

Driving safety is pursued from different perspectives by research communities, as shown in Fig. \ref{fig:topics}. 
New safety and warning systems are under design and development. Examples are applying eye-tracking technology to assess drivers' distraction and fatigue \cite{xu2018real,le2020evaluating}, and onboard collision avoidance systems to offer safety alerts and active brake upon detecting a danger \cite{ba2017crash,abdul2018modular,mossy_toyota}. 
Another research direction is \ar{exploring the casualty of incidents based on crash surrogate events (as the measures of accident proximity) from long-period naturalistic driving data\cite{gordon2011analysis,tarko2019measuring}. These works show that the expected number of crashes to occur during a specific time period is related 
to the number of observed surrogate events and crash-to-surrogate factors \cite{wu2012crashes}. A triggered event often is determined by a set of key parameters known as \textit{surrogate safety metrics} (SSM) designed for human-driven cars \cite{mahmud2017application} as well as AVs equipped with autonomous driving system (ADS) \cite{wishart2020driving,wang2021review}. We also provide a comprehensive review of SSM in the \ref{sec:safety-metrics}. }  





The community also has taken advantage of the recent developments in DL-based image/video processing that yield superior performance far beyond the conventional methods. \ar{DL methods have also enabled developing well-annotated volume datasets (e.g., highD dataset 2018) \cite{krajewski2018highd} that in turn led to developing even more powerful video processing methods for autonomous and safe driving applications, such as vehicle detection \cite{liu2021real,qin2021traffic}, plate recognition \cite{khazaee2020real,chen2019automatic}, traffic sign classification \cite{liu2021real,qin2021traffic}, lane detection 
\cite{neven2018towards,roberts2018dataset}, and abnormal driving detection from surveillance video \cite{nguyen2020anomaly,giannakeris2018speed}.}


Nowadays, safety-related events such as traffic congestion, red light violation, over speeding, unauthorized-vehicle stops on the highway shoulder, etc., can be detected, interpreted, and predicted by learning-based video analysis frameworks, such as \ar{generative adversarial network} (GAN)-based architectures~\cite{nguyen2020anomaly}, query-based approaches \cite{fu2019rekall}, 3D convolutional Networks \cite{zhou2016spatial}, and YOLO-family detectors \cite{doshi2020fast}.

To the best of our knowledge, most current methods favor the investigation of individual and independent crashes based on the extracted incident-level safety metrics and disjoint safety events. Therefore, they are not well-positioned to make relations between the crash distributions and the dynamics of the entire traffic flow.

\begin{figure}[htbp]
\begin{center}
\centerline{\includegraphics[width=0.5\textwidth]{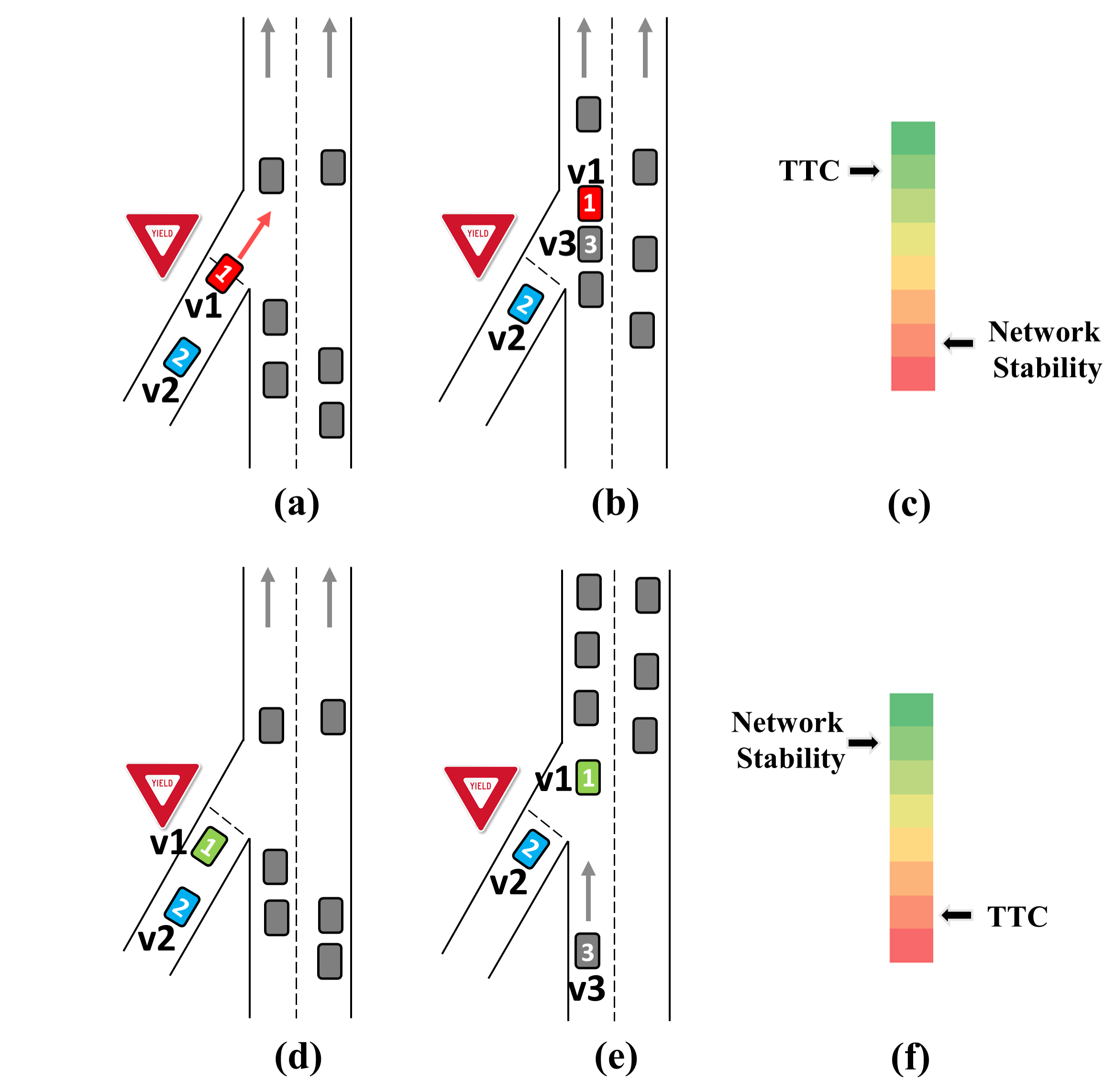}}
\caption{The individual analysis fails to interpret a merge scenario. Vehicle v1 on the entrance ramp intends to join the highway traffic. The top row (a,b) shows unsafe (aggressive) join before and after the merge. This is considered favorable by the following car (v2), since it provides a higher TTC, while disrupting the overall highway traffic stability (c). The bottom row shows the safe join by v1 after yielding the traffic flow before (d) and after the merge (e). Although this merge provides a lower TTC for the following vehicle on the ramp (v2), it is advantageous from the traffic stability point of view (f). This can be reflected in the TTC of the car following v1 after joining the highway (v3).}
\label{fig:comparison_2}
\end{center}
\end{figure}

\begin{figure*}[]
\begin{center}
\centerline{\includegraphics[width=0.9\textwidth]{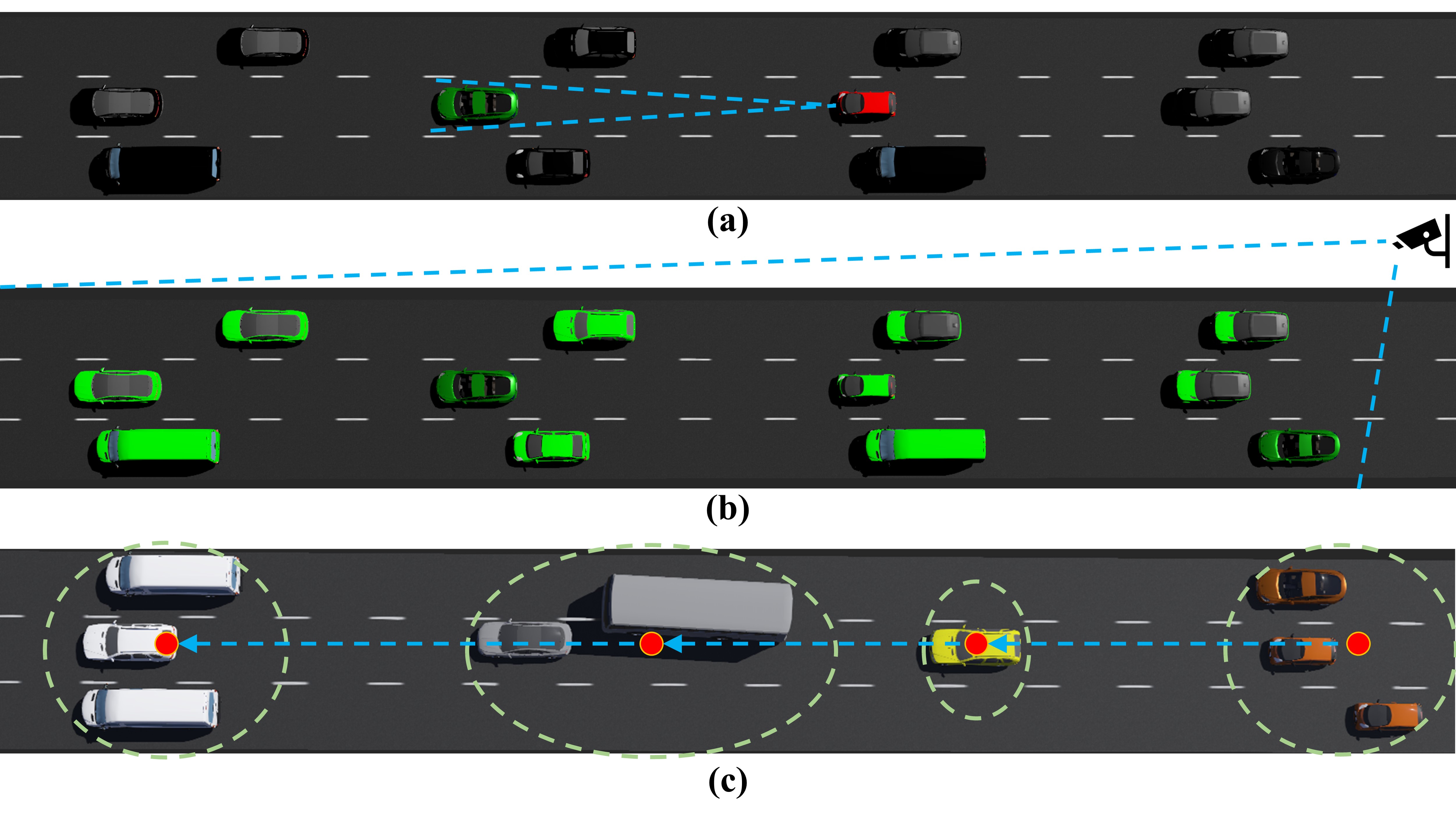}}
\caption{Different levels of traffic safety along with their regions of interest for (a) individual safety analysis from the vehicle's perspective using standard safety metrics, (b) global safety analysis from an external observer's perspective using the proposed network-level metrics, and (c) local analysis of traffic clusters.}
\label{fig:comparison_metrics}
\end{center}
\end{figure*}

Fig.~\ref{fig:comparison_2} demonstrates a merge scenario, where individual safety metrics may not be capable of capturing overall safety risks, hence can yield misleading results. 
Gray rectangles in this figure represent the normal traffic flow of the highway, and vehicle v1 intends to join the traffic. Figs. \ref{fig:comparison_2}(a) and \ref{fig:comparison_2}(b) present an aggressive merge before and after the joining epochs, while Figs. \ref{fig:comparison_2}(c) and \ref{fig:comparison_2}(d) show a safe merge. 
Let's investigate this scenario using the time-to-collision (TTC) metrics, which is one of the most commonly used safety metrics for safety analysis, especially rear-end crashes~\cite{ge2019construction,dimitriou2018assessing}. 
Calculating TTC for the leading and following cars (v1 and v2) will favor the aggressive join, since a faster merge leaves more reaction time for the following car v2, and hence appears safer from v2's perspective. However, it causes more risk to the following vehicle on the highway after joining the traffic (v3 in Fig. \ref{fig:comparison_2}), hence disrupting the stability of the traffic flow. This simple scenario shows how individual two-car investigations can lead to misleading results. Indeed, traffic flow can represent a complex dynamic system with many factors interacting with one another, requiring an overall network-level analysis. For instance, a traffic blockage in one intersection can influence the traffic volume (and hence the traffic safety) of alternative nearby routes.  

It is noteworthy that simply extending the results of individual crash analyses does not fully address this issue since the safety metrics are calculated only for vehicles that involve in crashes, while our perspective is profiling the overall traffic safety through network-level metrics.

\begin{figure*}[]
\begin{center}
\centerline{\includegraphics[width=1\textwidth]{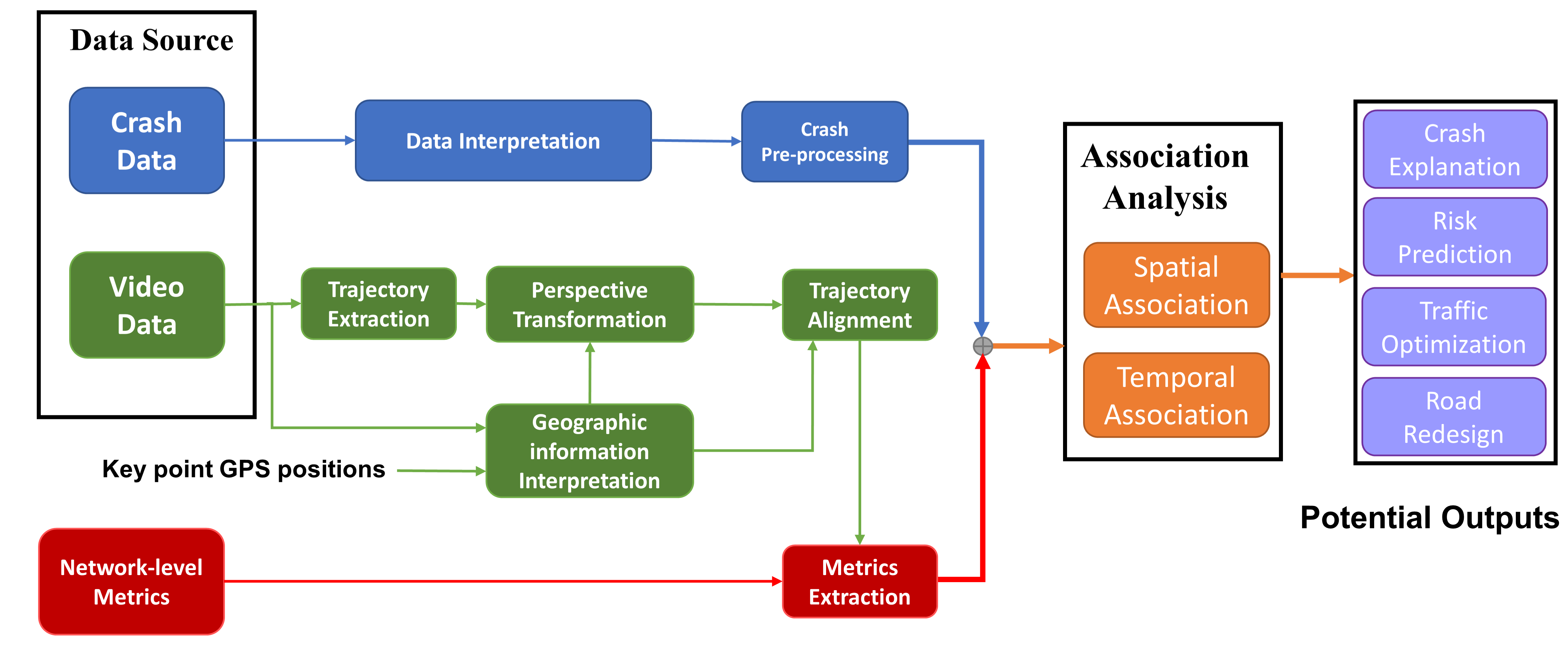}}
\caption{The proposed network-level traffic safety analysis framework includes multiple modules: (i) video processing that includes video preparation (e.g., video stabilization, noise reduction, and personal information masking), trajectory extraction, geo-mapping with perspective translation, and network-level safety metric extraction, (ii) crash analysis which includes crash temporal and spatial mapping, crash distribution (histogram), and crash type mapping (representation), and  (iii) temporal and spatial association analysis. 
}
\label{fig:pipline}
\end{center}
\end{figure*}

\ar{One may argue that one easy solution would be averaging individual safety metrics among all vehicles. However, we show that elegantly designed network-level safety metrics can provide further information about safety risks. For instance, we conjecture that the spatial distribution of vehicles on the road correlates with the crash probability. In this respect, we group vehicles into clusters and calculate cluster-wise TTC metrics (TTC-CV) as elaborated in section \ref{sec:nms}. Then, we find the correlation between the TTC metrics and crash rates. As shown in Table \ref{tab:simple_TTC}, the proposed TTC-CV metrics show three times higher correlations with crash rates compared to that of the individual TTCs averaged over all vehicles, E(TTC). This is consistently true for all crash types and rear-end crashes using three measurable correlation methods, Pearson, Spearman's, and Kendall.}




\begin{table}[]
\centering
\caption{\ar{Comparison of the correlation to crashes with a simple average value of TTC and our proposed TTC-CV.}}
\label{tab:simple_TTC}
\resizebox{0.45\textwidth}{!}{%
\blt\begin{tabular}{lllllll} \toprule
         & \multicolumn{2}{c}{Pearson}       & \multicolumn{2}{c}{Spearsman's}   & \multicolumn{2}{c}{Kendall}       \\ \midrule
Metrics  & \textbf{TTC-CV} & \textbf{E(TTC)} & \textbf{TTC-CV} & \textbf{E(TTC)} & \textbf{TTC CV} & \textbf{E(TTC)} \\
All Type & 0.308           & -0.014           & 0.384           & -0.014           & 0.273           & 0.086
           \\
Rear-end & 0.310           & -0.004          & 0.373           & -0.004          & 0.265           & 0.082
     \\ \bottomrule     
\end{tabular}%
}
\end{table}

The importance of macro-scale analysis of crash data has been recognized by some researchers \cite{harirforoush2019spatial,al2021mapping,wang2020random,meng2018expressway,sipos2021spatial}. However, most of these studies investigate the geo-spatial distribution of crashes and their consequences with limited insight to finding diverse causes of crashes. More specifically, they emphasize geographical mapping of traffic properties (e.g., volume, density, congestion condition, etc.) as well as the road network topology. 
The high-resolution micro-level driving behaviors, such as multi-agent trajectory prediction~\cite{mozaffari2020deep}, motion planning~\cite{claussmann2019review} have also been studied. These analyses predict crash probabilities and safety factors but they didn't try verifying their results with readily available crash reports. 

A brief review of these methods is provided in Section \ref{sec:net_analysis}. 
\ar{It is noteworthy that our analysis nicely integrates with these studies, since these methods find the baseline crash probability for each road section, whereas our method quantifies the modulated crash probability (the amount of increase/decrease) based on the temporal safety profiling of the traffic flow.}

In this work, we offer a \ar{generalizable} framework for finding meaningful correlations between representative traffic safety indicators and crash probability geo-distribution through integrative analysis of crash reports and traffic video captured by the roadside infrastructure. Our contribution is two-fold. First, we define a set of network-level safety metrics (NSM) from an external observer's perspective that captures the inherent relations between local traffic flows and gauge the overall safety profile of the traffic network (Fig. \ref{fig:comparison_metrics}(b)). These metrics are not claimed to be comprehensive, and are presented only to show the utility of such metrics for traffic analysis as a proof-of-concept, and can be extended to a more comprehensive list. 
Secondly, we provide a formal association analysis to assess the contribution of each safety metric in mediating the crash frequency. This is done by investigating the spatial and temporal correlation between the crash data points and the traffic disruption represented by the proposed safety metrics. The results of our framework can be used for developing online traffic advisory systems, crash explanation, risk prediction, and traffic optimization. \ar{It is noteworthy that crash severity analysis is another legitimate problem which is out of the scope of this paper. We have to acknowledge that our metrics may be not directly used to analyze crash severity since crash severity is more heterogeneous by nature and involves more confound factors beyond the developed NSMs.}  

The overall pipeline of the proposed integrative framework is presented in Fig. \ref{fig:pipline}, which includes two parallel processing modules for crash analysis and video-based safety metric extraction followed by an association analysis step. 

\section{Related Work on Network-level Analysis}\label{sec:net_analysis}

Traffic analysis can be performed at different levels. In micro-level analysis, typically an individual incident is analyzed by a set of fine-resolution parameters such as the involving vehicles' geometry and motion dynamics as well as the local road and traffic flow properties, based on the sensor readings and captured imagery. A comprehensive review of micro-level analysis of crashes, DL-based driving behavior modeling, AI-platforms for driving safety enhancement have been reviewed in~\cite{zheng2021modeling,mozaffari2020deep,claussmann2019review}.

In macro-level analysis, the traffic flow is considered a dynamic network with inherent relations among network nodes. This view is adopted for network structural analysis, traffic forecasting, abnormal pattern detection, and global traffic safety analysis. 
For instance, \cite{wu2004urban,gao2007study,boccaletti2014structure,klinkhamer2017functionally} consider traffic as an application of complex network theory, where the network dynamics can be represented by the collection of small-world networks \cite{watts1998collective} and random scale-free networks \cite{barabasi1999emergence}. A small-world network is defined as a network constructed with a high clustering coefficient with small average geodesics, namely the pairwise shortest path lengths. In other words, most network nodes are expected to be accessible with a low number of hops, (e.g., logarithmic in the number of nodes). This model best fits a collection of lattice-shaped local neighborhood roads linked with a few highways.

A scale-free network is a network whose degree distribution obeys the Power-law, meaning that there exist only a few nodes with higher degrees (such as downtown or traffic hubs). Urban traffic can be modeled by such networks \cite{,wu2004urban}.

Some other works \cite{batty2012building,batty2013new,boccaletti2014structure,ding2018detecting,wang2017improved,ruan2019empirical,klinkhamer2017functionally} analyze the network structure using complexity networks theory to evaluate, design and optimize traffic networks with sustainability and maintainability.
Detecting network bottlenecks with poor connectivity and high congestion vulnerability is studied in \cite{newell1998moving,hasan2012modeling,mccrea2010hybrid,long2008urban,nakata2010dilemma,li2020congestion,qu2019road}. \ar{The authors of \cite{mohammadian2021performance} provide a comprehensive review of continuum models that consider the traffic flow as a compressible fluid and employ some physical knowledge to explain the real-world phenomena in traffic.} 
It is noteworthy that some recent works \cite{lu2020impact,parsa2020data,mavromatis2020urban} address mixed traffic taking autonomous vehicles (AVs) into consideration in their analysis.

Traffic safety can be improved by more accurate traffic forecasting, namely by predicting the future properties of the traffic flow based on current/historical features. Traffic risk modeling and prediction can provide hints and guidelines on traffic management to minimize factors that elevate the crash risk.
Specifically, \cite{zhang2016dnn,zhang2017deep} transform the road network to 2D grids and then apply convolutional neural network (CNN) to predict the traffic crowd flow. The traffic flow is modeled as a diffusion process on a directed graph and deploys a diffusion convolutional recurrent neural network (DCRNN) to learn the spatio-temporal features of traffic based on the historical data and road structure \cite{li2017diffusion}. A method called Deep Transport is proposed in \cite{cheng2018deeptransport} which combines CNN and recurrent neural network (RNN) architectures equipped with an attention mechanism to predict traffic volume. Recently, graph neural networks (GNNs), a class of DL networks performing inference over arbitrary graphs are proven to yield superior performance in predicting traffic-related parameters \cite{yu2017spatio,yu2019real,yu2020forecasting,fu2020bayesian,jin2020urban}.

Crashes can also be viewed as severe disruptions in the network flow. Therefore, many papers have focused on abnormal network pattern detection, such as inferring abnormal patterns caused by unexpected events (e.g., natural disasters, serious car accidents, traffic control) \cite{zhang2016spatial,zhang2017abnormal,huang2018traffic}.
For instance, interpreting crashes by analyzing the frequency of irregular patterns over traffic networks is considered in the following works. 
The role of road factors in characterizing the severity of incidents using logistic regression with chi-squared test is presented in \cite{pugachev2017factor}.  
Kernel density estimation (KDE) is used in \cite{harirforoush2019spatial,al2021mapping} to find the spatio-temporal patterns of traffic accidents, and rank them based on their statistical significance. 
Negative binomial and Poisson models are used in \cite{wang2020random} to identify traffic factors that contribute to the crash frequency. Crash prediction based on random forest (RF), gradient boosting decision tree (GBDT), and Xgboost is performed in \cite{meng2018expressway}. A spatial econometric model is adopted to find associations between the road links and incidents \cite{sipos2021spatial}.

The above-mentioned works provide insightful results for different traffic problems by network-level analysis of traffic flow and crash statistics. However, they still suffer from a few drawbacks. Some works are devoted to explaining the traffic flow and crash distributions using deep learning models. Although successful from modeling and prediction perspectives, these methods lack the interpretability and generalizability features. Moreover, due to the difficulty of creating crash scenarios, the micro-level analysis is hardly verifiable, and most works settle with verifying the results with virtual simulators such as car learning to act (CARLA) \cite{Dosovitskiy17} and simulation of urban mobility (SUMO) \cite{sumo}, ignoring the valuable information exploitable from a rich set of well-documented crash data. Furthermore, the utilized safety metrics are appropriate only for individual crash analysis, hence fail in modeling the inherent and complex relations among network nodes. In this work, we make a connection between the global analysis and deep analysis of individual incidents by introducing newly-defined network-level metrics. The integrative analysis of traffic video and crash data enables us to look for \ar{statistical} 
relations between traffic properties and crash risks and drawing general conclusions on traffic safety enhancement. \ar{It is notable that there is a gap between the association analysis using data-driven methods and the real causality identification studied in \cite{mannering2020big}.} 

\ar{Finally, we note that our method does not replace the micro-level and macro-level analyses, but complements them. It extends the notion of safety metrics to more insightful network-level metrics for global crash analysis while profiling the overall traffic flow safety that can fine-tune the baseline crash probability obtainable from macro-scale analysis of crash reports.} 



\section{Proposed Metrics for Network-level Analysis}  \label{sec:nms}
\ar{Conventional safety metrics are defined for scenarios where only two (or a few) vehicles are involved.}  \ifx \includeAppendix \incYes 
A summary of the most commonly used safety metrics is provided in Appendix A, for the sake of completeness. In this appendix, we also introduce a taxonomy for safety metrics based on the level of access required to ADS data, when calculating these metrics for self-driving cars.
\fi
In this paper, we introduce a set of NMS that can be used for the overall and long-term safety assessment of traffic flow. 
For instance, the composition of traffic (e.g., the ratio of trucks to all vehicles) can contribute to the frequency and severity of collisions. Likewise, the overall variations of car velocities on the road can reveal information about potential risk factors. 
A list of the proposed network-level safety metrics is provided in Table \ref{tab:newmetric}.

\textbf{Time-to-Collision-Cluster Variation (TTC-CV)} is defined to evaluate the relative velocity of car clusters that can pose safety risks. TTC is perhaps the most commonly used safety metric that evaluates the risk of read-end crash by quantifying the time of the following car colliding with the leading car if they both retain their current speeds. We develop a new metric that extends this metric to car clusters \ar{since vehicle clustering naturally occurs on the road. We also conjecture that the instantaneous geo-distribution of the cars on the road can play essential roles on crash rates. For instance, car platooning is considered an important feature for autonomous and connected vehicles. For such scenarios, due to the coordination between platooning vehicles, inter-platoon crashes are rare and cluster-based TTCs can be useful.}

Our approach is clustering vehicles based on a predefined threshold and assess the relative mobility of car clusters. 
More specifically, we use the down-sampled version of the traffic video, e.g., at rate 1 FPS) to cluster the cars based on their pairwise distances. 
If $\mathcal{S}$ is the set of cars in the current video frame, then a cluster $C_i =\{n_j\}\subset {S}$ is defined so that for every $n_j \in C_i$, there exists at least one node $n_k \in C_i,~j \neq k$ with $d(n_i,n_j)\leq d_{C}$ and likewise any car $n_l$ satisfying $d(n_k,n_j)\leq d_{C}$ for some $n_j \in C_i$ must be a member of $C_i$. Here, $d(n_i,n_j)$ is the Euclidean distance between vehicles $n_i$ and $n_j$, and $d_C$ is a predefined threshold. The clusters are non-overlapping and we have $C_i \cap C_j=\{\}$ for all $i \neq j$. 
Then, the cluster $C_i$ is considered as a virtual point object at the centroid of the cluster, i.e. $l(C_i)=\sum_{n \in C_i} l(n)/|C_i|$ with velocity $v(C_i)=\sum_{ n\in {C_i}}{v(n)}/|C_i|$. Here, $l(n)$, $v(n)$
represent the location and velocity of node $n$, and, $|C|$ is the cardinality of set $C$ (i.e. the number of cars in cluster $C$). The TTC of $C_i$ is calculated with respect to the potential collision point with the latest leading cluster $C_j$ moving at a lower speed $v(C_j) \leq v(C_i)$ as follows:

\begin{align}\label{eq:ttci}
    CTTC_{i} = \frac{l(C_j)-l(C_i)}{v(C_i)-v(C_j)}.
\end{align}

For segments with crossing road segments (intersections and merging points), we use the stationary intersection point as the potential collision point when calculating cluster TTCs. Also, note that cluster TTCs are calculated using the original video with high FPS 30. 
The cluster TTC reduces to the regular inter-vehicle TTCs if the threshold $d_C$ is selected close to 0, so each car becomes a cluster. 
Next, the coefficient of variation of cluster-level TTCs for frame $j$, is calculated as:

\begin{align}
TTC\text{-}CV(f_j) = \frac{\text{std}(CTTC)}{\text{mean}(CTTC)} \rho_v,       
\end{align}
as an instantaneous network-level collision risk factor for the road segment covered by video frame $f_j$. Here, $N_v$ and $N_C$ are the numbers of vehicles and clusters in the frame, and $\rho_v={N_v}/{N_c}$ is the average number of vehicles in each cluster used to emphasize higher risk for more crowded clusters. \ar{This metric is more robust against outliers and extremums (which often occur in computing TTC) compared to other statistics of cluster-level TTCs, including $\min(CTTC)$,  $\text{mean(CTTC)}$, or $\max(CTTC)$.}

\textbf{Individual Velocity Variation Rate (IVVR)} is defined as the variation of velocities for each vehicle in a specific zone or time interval. We define it as

\begin{align}
    IVVR =\frac{1}{N}\sum_{i=1}^{N} \frac{|v_i^\text{max}-v_i^\text{min}|}{v_i^\text{av}},
\end{align}
where $v_i^\text{max},v_i^\text{min},v_i^\text{av}$ are the minimum, maximum, and average velocities of vehicle $i$. The higher values of this metric means that on average each vehicle changes its speed more often by accelerating and decelerating. This can be due to the road profile, density of intersections and exits, traffic volume, or road conditions.

\begin{table*}[htbp]

\centering
\resizebox{0.8\textwidth}{!}{%
\begin{tabular}{lll}
\toprule
\textbf{Metric} &
  \textbf{Definition} &
  \textbf{Features} \\ \midrule
  TTC-CV &
  $TTC\text{-}CV(f_j) = \frac{\text{std}(CTTC)}{\text{mean}(CTTC)} \rho_v$ &
  \begin{tabular}[c]{@{}l@{}} Extend the conventional TTC to cluster level; \\ Evaluate the global risk of the traffic flow at the network-level.\end{tabular} \\ \midrule

IVVR &
  $IVVR =\frac{1}{N}\sum_{i=1}^{N} \frac{|v_i^\text{max}-v_i^\text{min}|}{v_i^\text{av}}$ &
  \begin{tabular}[c]{@{}l@{}} Reflects the instability of traffic flow by car speed variations and crash rate;\\ Can partially offer network re-design suggestions\end{tabular} \\ \midrule
OVVR &
  $OVVR =\frac{1}{N}\sum_{i=1}^{N} \frac{|v_i^\text{av}-v^\text{av}|}{v^\text{av}}$ &
  \begin{tabular}[c]{@{}l@{}}Overall speed variation of vehicles; \\Defined similar to IVVR for overall traffic speed variation;\\ Not easily affected by outliers \end{tabular}\\ \midrule
OSR &
  $ OSR_i =\frac{1}{N}\sum_{i=1}^{N} I(v_i^\text{max}/v_L > Threshold_i)$ &
  \begin{tabular}[c]{@{}l@{}}Indication of speed limit violations; Can offer suggestions for network design\end{tabular} \\ \midrule
TCI &
  \begin{tabular}[c]{@{}l@{}}$TCI = \frac{\big(\sum_{c=1}^C N_c)^2}{C\sum_{c=1}^C N_c^2}$\\ $f_c = \frac{N_c}{\sum_{c=1}^C N_c}$\\ $TCI = \frac{1}{2(1-2 f_1 f_2)}$, two classes\end{tabular} &
  \begin{tabular}[c]{@{}l@{}}Flow composition indicator;\\ Can be associated with elevated crash risks\end{tabular} \\ \midrule
NTC &
  $NTC = \frac{\sum l_i}{N_l*L}$ &
  \begin{tabular}[c]{@{}l@{}}Consider vehicle shape;\\ Reflects traffic density;\\ Can be associated with elevated crash risks\end{tabular} \\ \midrule
TRT &
  $TRT = \frac{1}{N_e}\sum_{i=1}^{N_e} |t_r(i)-t_b(i)|$ &
  \begin{tabular}[c]{@{}l@{}}
  A simple and reasonable way to evaluate severity of the accidents;\\
  Can also take traffic stability into consideration\end{tabular}\\ \bottomrule
\end{tabular}%
}
\caption{A set of proposed network-level safety metrics.} 
\label{tab:newmetric}
\end{table*}

\textbf{Overall Velocity Variation Rate (OVVR)} is defined as the variation of average velocities among vehicles in a specific zone or time interval. OVVR is defined as 

\begin{align}
    OVVR =\frac{1}{N}\sum_{i=1}^{N} \frac{|v_i^\text{av}-v^\text{av}|}{v^\text{av}},
\end{align}
where $v_i^\text{av}$ is the average velocity of vehicle $i$, and $v^\text{av}=\sum_{i=1}^N{v_i^\text{av}}/N$ is the average velocity of all vehicles. Similar to IVVR, this metric can be associated with crash rate in specific highway sections.

\textbf{Over Speeding Rate (OSR)} is defined as the rate of over-speeding vehicles as follows:

\begin{align}
    OSR =\frac{1}{N}\sum_{i=1}^{N} I(v_i^\text{max} > v_L),
\end{align}
where $v_L$ is the speed limit, and $I(x>y)$ is the indicator function with $I(x>y)=1$ for $x>y$ and $I(x>y)=0$ otherwise.   
Speed limits are typically set based on a standardized set of national guidelines, taking into account road geometry (e.g., radii of curves, sight distance, weather conditions) and the location profile (e.g., residential versus rural areas). A high OSR, when associated with a high crash rate, may indicate the need for taking more warning and prevention measures to avoid over-speeding, noting that over-speeding can be an important contributor to crashes. On the other hand, high OSR, when it does not correlate with a high accident rate, can indicate that speed limits could be considered for potential increase without compromising traffic safety. 

The aforementioned traffic metrics can be characterized by processing the roadside cameras or by crowd-sourcing and accumulating information provided by vehicles' dash cameras.

\textbf{Traffic Composition Indicator (TCI)} is defined to gauge the diversity of vehicle types in specific road sections.  
For instance, it is known that a higher density of trucks on the roads can correlate with the frequency and severity of road accidents; hence trucks are prohibited in some road sections in highly-populated areas \cite{meyers1981comparison}. This can be due to trucks' larger deceleration inertia, more frequent break failures, and lower maneuverability levels. In general, if vehicles classified into classes $c=1,2,3,\dots,C$, then the traffic composition can be defined as

\begin{align}
    TCI = \frac{\big(\sum_{c=1}^C N_c)^2}{C\sum_{c=1}^C N_c^2},
\end{align}
where $N_c$ is the number of vehicles in class $c$, and the metric is defined similarly to the Jain fairness index. This metric ranges from $1/C$ for the most unbalanced composition to $1$ for an equal number of vehicles of each type. If one is interested in evaluating the rate of a specific class like trucks to all vehicles, the following metric can be used:

\begin{align}
     f_c = \frac{N_c}{\sum_{i=1}^C N_i}.
\end{align}

When classifying cars into two classes $c=1$ for trucks and $c=2$ for non-trucks, these two metrics are related as 

\begin{align}
     TCI = \frac{1}{2(1-2 f_1 f_2)}.
\end{align}

Lower $TCI$ values mean that most cars are of the same type with lower risks. This metric can be estimated by processing roadside videos but requires high-complexity learning methods for vehicle detection and classification. 
 
\textbf{Normalized Traffic Density (NTC)} is defined as the density of vehicles on road sections as 
 
  \begin{align}
     NTC = \frac{\sum l_i}{N_l*L},
 \end{align}
where $l_i$ is the length of vehicle $n_i$, $N_l$ is the number of lanes, and $L$ is the length of the road section. We normalize the vehicle length to capture the effect of vehicle's size. 
This parameter can be easily extracted from roadside videos by video processing and vehicle detection. Higher NTC values are expected to be associated with higher crash rates, and may raise the request for traffic load balance strategies.  
 
\textbf{Traffic Recovery Time (TRT)} is defined as the time required for traffic flow recovery after incidents. 
The system of vehicular flow can be considered as a non-equilibrium system of interacting particles \cite{sugiyama2008traffic}, and the instability of a flow-free state is induced by the collective effects of the increase of fluctuations. 

\begin{figure}[htbp]
\begin{center}
\centerline{\includegraphics[width=0.8\columnwidth]{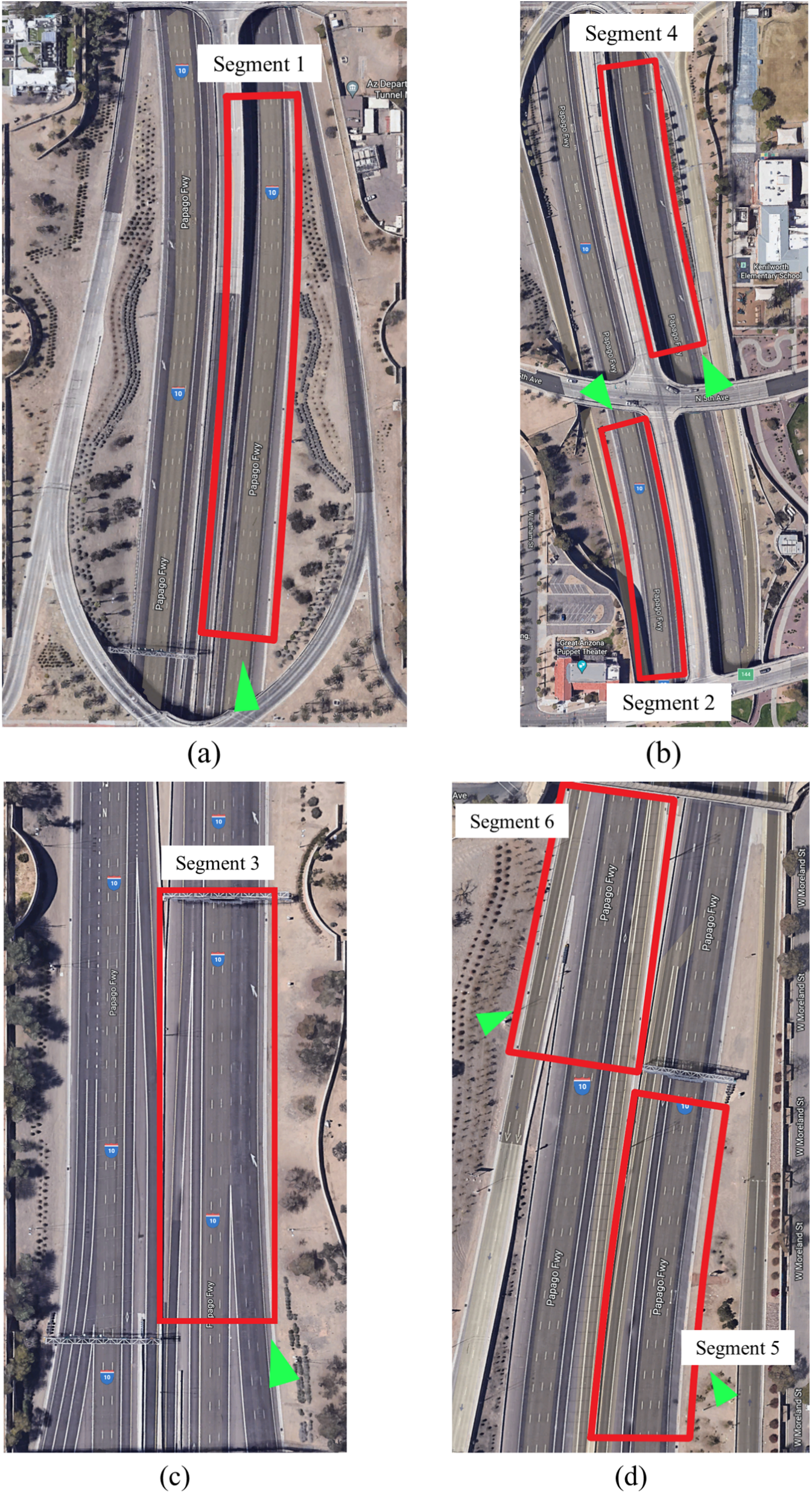}}
\caption{\ar{Six road segments (equivalently six selected areas for perspective transformation) are shown by red squares, and the camera locations are shown by green triangles.}}\label{fig:segment_123456}
\end{center}
\end{figure}

\begin{figure*}[h]
\begin{center}
\centerline{\includegraphics[width=0.9\textwidth]{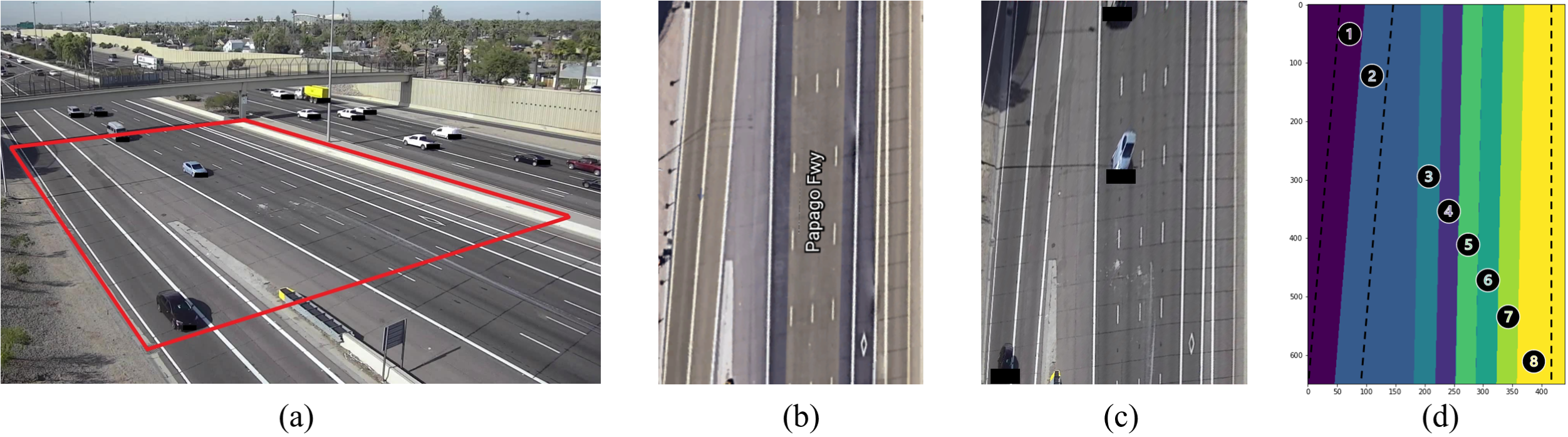}}
\caption{\ar{Perspective transformation for fixed camera view of Segment 6: (a) Selected area covered by an RSU camera field of view, (b) GPS view of the selected area, (c) aligned perspective of the selected area, and (d) lane segmentation of selected area.}}\label{fig:lane_mask}
\end{center}
\end{figure*}

It is known that any traffic event can suddenly lead to a jamming state, and TRT reflects the time span from an event epoch to the time point the status changes to free flow. This time is expected to be much shorter than the interval between the consecutive events for flow stability. TRT can be defined as:

\begin{align}
    TRT = \frac{1}{N_e}\sum_{i=1}^{N_e} |t_r(i)-t_b(i)|,
\end{align}
where $N_e$ is the number of events in the monitoring interval, and $t_b(i)$ and $t_r(i)$, respectively, denote the event start epoch, and the flow recovery epoch for event $i$. This parameter can be easily obtained from roadside videos or by crowd-sourcing the position information obtained from dash cameras.

The summary of the proposed network-level safety metrics is presented in Table \ref{tab:newmetric}. We believe that further efforts are needed to develop a more complete list of network-level safety metrics.

\section{Methods}
\subsection{Data acquisition}
In this work, we use \ar{six} camera feeds collected by the Arizona Department of Transportation (ADOT) roadside infrastructure.
Each camera covers one segment of highway I-10 and records five 2-hour MP4 videos with resolution 1280x720 and FPS 30 (each video file is about 8 Gigabytes). \ar{Six} exemplary covered highway segments along with the approximate camera locations are shown in Fig.~\ref{fig:segment_123456}.

\subsection{Video preprocessing}\label{sec:vid_process}

In order to calculate safety metrics, we need to extract motion trajectories of vehicles from the video files. To this end, we integrate a tracking algorithm called \textit{simple online and realtime tracking with deep association metric} (DeepSORT) \cite{wojke2017simple} with the \textit{you only look once} (YOLO)-v5 \cite{glenn_jocher_2021_4679653} detector to obtain trajectories of labeled objects. DeepSORT is a real-time multi-object tracking algorithm based on Kalman filtering and Hungarian algorithm, which can consider both bounding box parameters and appearance simultaneously. Known for its speed and accuracy, YOLOv5 is an advanced proposal-free detector adopted to sustain the high-quality achievement of tracking. The centroid of detected bounding boxes is used as the position of the objects.  
This combination offers real-time tracking ($\sim$40 FPS with a GeForce RTX 2070 GPU) with acceptable accuracy (>90\%). \ar{This setup is sufficient to perform experiments on certain road segments. To the best of our knowledge, as the computer science and industry thrive, more lightening and accurate algorithms with hardware acceleration would arise that could allow widespread deployment of our framework on RSU of large-scale road networks.} 
Another advantage of this approach is tracking objects even with long occlusion periods, a frequent issue in multi-vehicle tracking.


The extracted trajectories are in the pixel domain from the camera's perspective, hence the exploited distances and velocities are not proportional to real values. In order to extract safety metrics from trajectories, we translate the position information ($u,v$) from the 2D pixel domain into 3D GPS positions ($x,y,z$) using perspective projection, \ar{as shown in Fig. \ref{fig:segment_123456} and Fig. \ref{fig:lane_mask}}.
To this end, we solve the following projection equations for a set of key points with known GPS positions.
Considering a flat surface with no elevation change, we can skip $z$ in our calculations.

\begin{equation}
\left(\begin{array}{l}
x \\
y \\
z \\
\lambda

\end{array}\right)=\left(\begin{array}{llll}
a_{11} & a_{12} & a_{13} \\
a_{21} & a_{22} & a_{23} \\
a_{31} & a_{32} & a_{33} \\
a_{41} & a_{42} & a_{43}
\end{array}\right)\left(\begin{array}{l}
u \\
v \\
1
\end{array}\right),
\end{equation}
where, $(x/\lambda,y/\lambda,z/\lambda)$ denotes the GPS positions of the pixel after transformation from the pixel index values $(u,v)$ with $\lambda$ being the scale factor. The optimal transformation coefficients $a_{11}, ..., a_{43}$ are obtained by applying the least squares estimation (LSE) to the set of selected keypoint pairs with known GPS positions. Under linear transformation (no camera edge distortion) four keypoints are sufficient to recover the projection matrix, but a higher accuracy can be achieved using more key points.

\begin{figure}[]
\begin{center}
\centerline{\includegraphics[width=0.8\columnwidth]{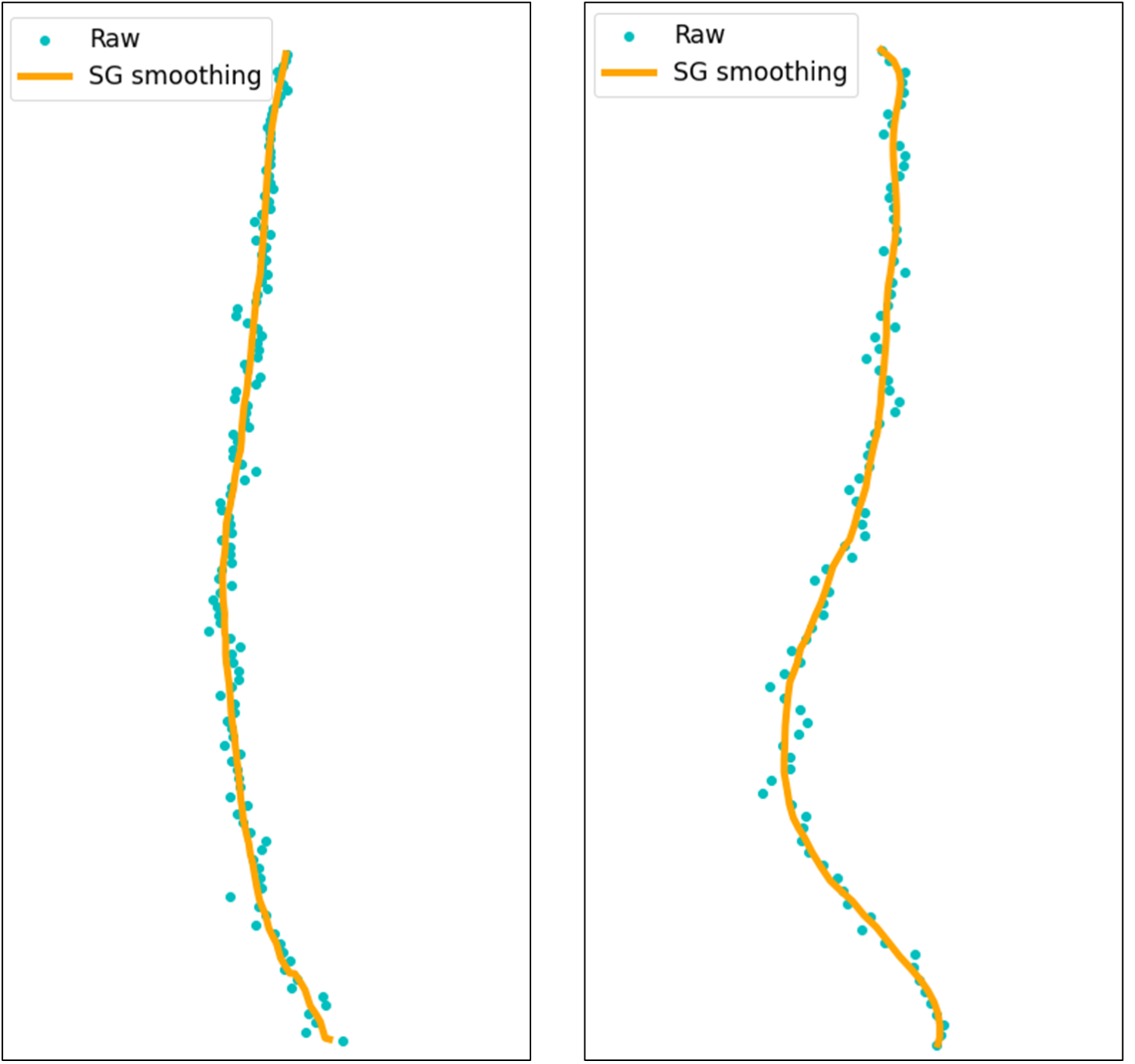}}
\caption{Some smoothed trajectories by SG filter and their original form.}\label{fig:SG-traj}
\end{center}
\end{figure}

The obtained trajectories represent noise-like fluctuations mainly due to the drift in the bounding boxes position around the object. We utilize a 
Savitzky–Golay (SG) filter \cite{schafer2011savitzky} to smooth out the trajectories before performing the subsequent association analysis (Fig.~\ref{fig:SG-traj}).

\subsection{Crash data}

\begin{figure}[htbp]
\begin{center}
\centerline{\includegraphics[width=1\columnwidth]{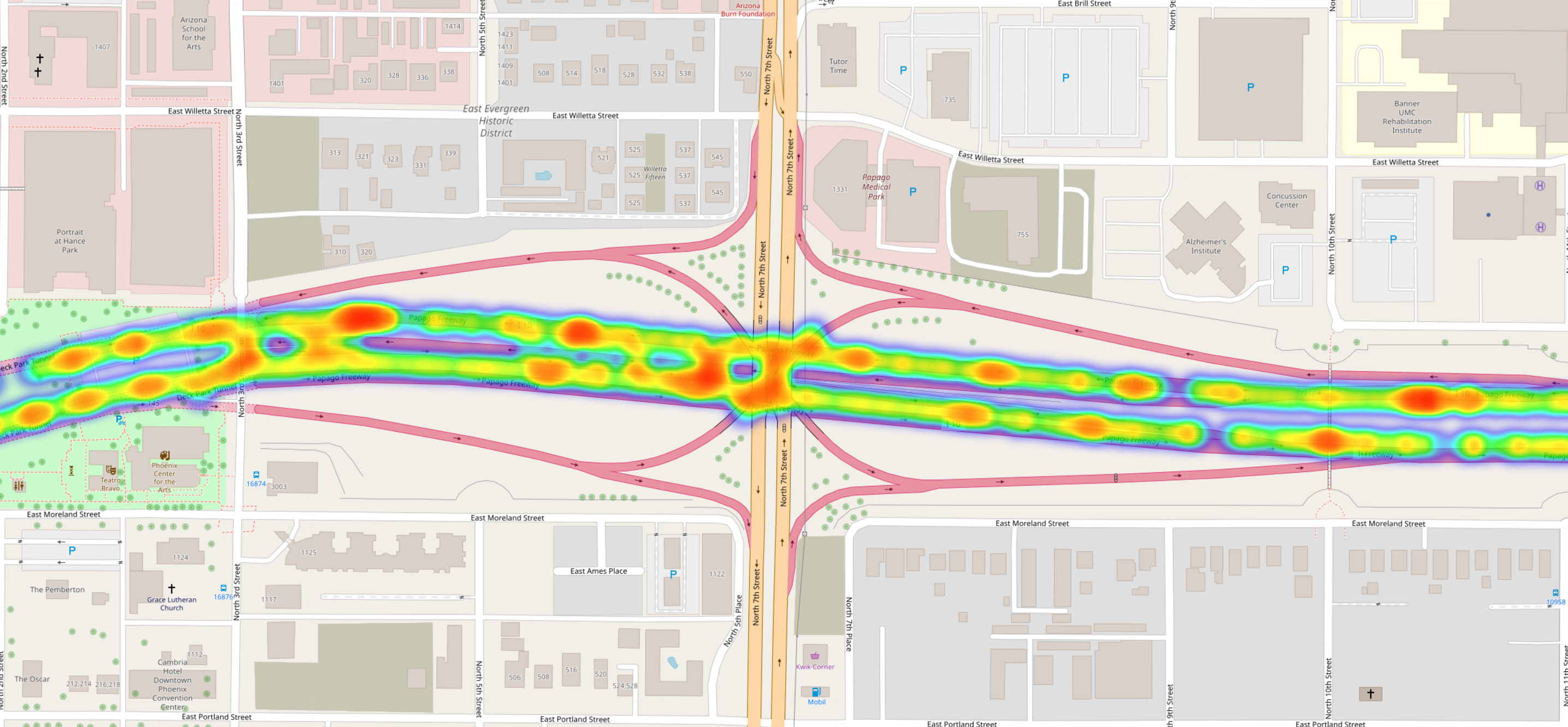}}
\caption{Geo-distribution of crash data is represented as a heatmap.}\label{fig:crash_heatmap}
\end{center}
\end{figure}

\begin{figure}[htbp]
\begin{center}
\centerline{\includegraphics[width=1\columnwidth]{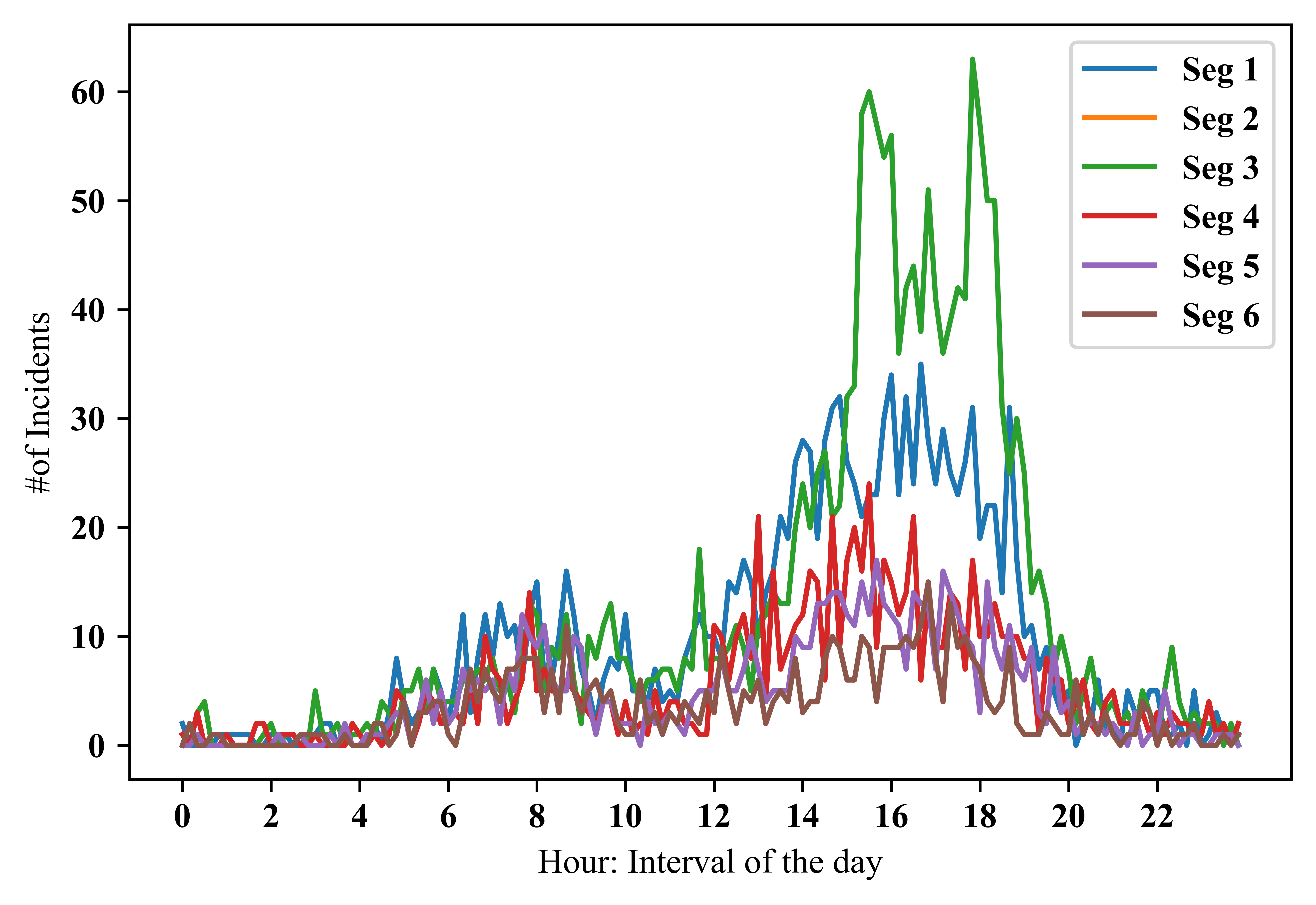}}
\caption{\ar{Crash count for six segments. The temporal trends for segments are consistent.}} \label{fig:viewcrash}
\end{center}
\end{figure}

The crash data is also provided by ADOT, which includes crash incident details in terms of date, time, location, collision type, etc. for 5-years, from 2015 to 2019. we extract the data for covered segments and use the most dominant crash types, including the rear-end, side-swipe, and all-types, since other types like head-on, angle, and rear-to-side crashes are sparse with not enough samples for association analysis. Figs. \ref{fig:crash_heatmap}, and \ref{fig:viewcrash} represent the geo-distribution of crashes and the crash statistics of the six segments, respectively. 

\ar{The temporal analysis aims to verify the utility of the proposed metrics in predicting crash count in each road segment for a given time interval. The spatial analysis investigates the generalization of the identified relations to other road segments with similar conditions. 
To this end, we split time into 1-hour intervals. For each interval, we calculate the average of safety metrics extracted from the traffic video for a specific road segment. Likewise, we take the average of crash counts for the same interval and road segment over the 5-year period. The validity of association analysis relies on two assumptions (i) different time intervals (e.g., [8:00 am-9:00 am] and [12:00 pm-1:00 pm]) are statistically different, and (ii) the crash counts over the same time intervals are statistically identical across different dates.}

\begin{figure}[htbp]
\begin{center}
\centerline{\includegraphics[width=0.6\columnwidth]{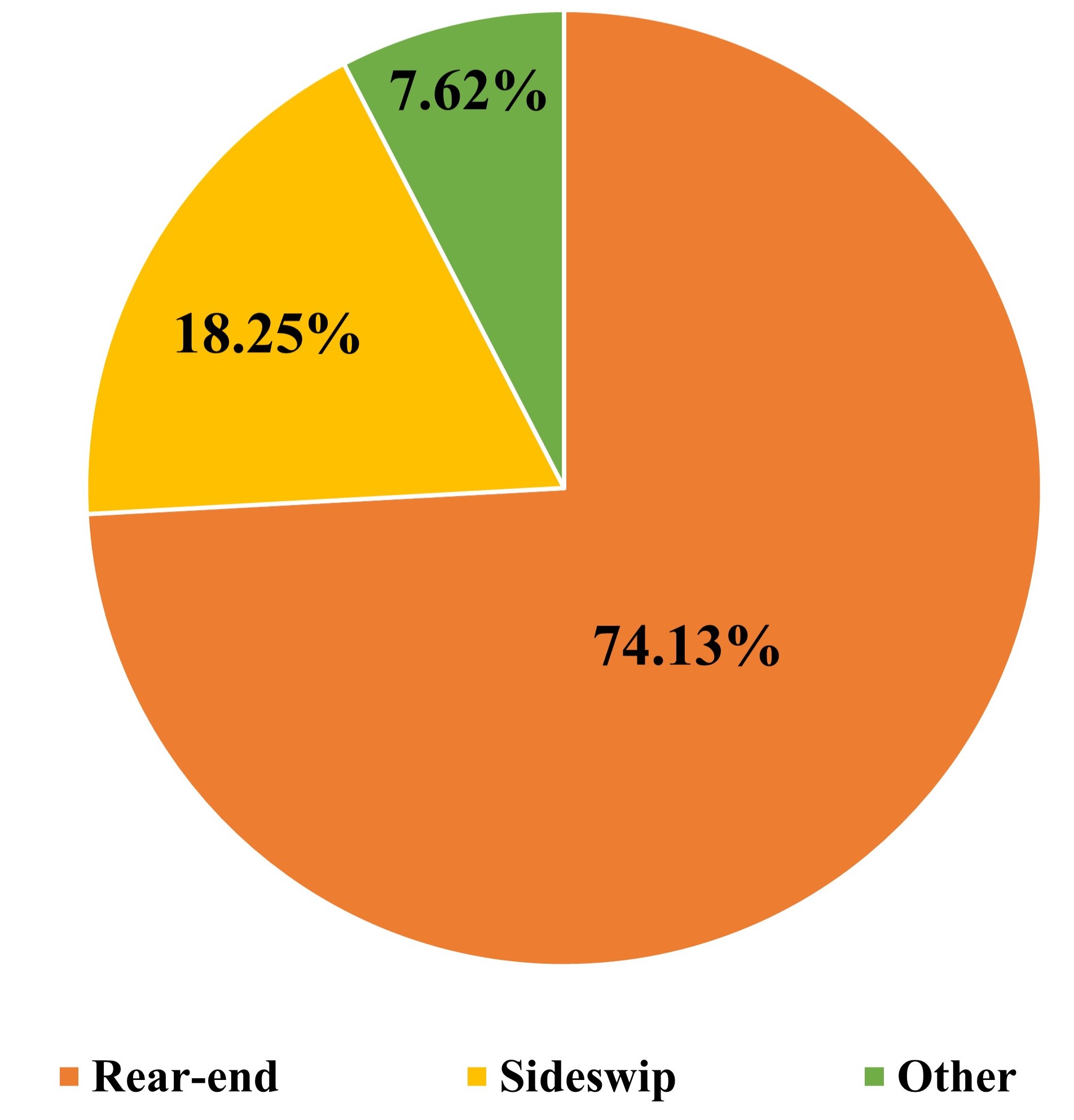}}
\caption{Crash types.}\label{fig:viewcrash_propotion}
\end{center}
\end{figure}

\ar{The results in Table \ref{tab:select_vs_entire_1} present the \textit{contingency Chi-square} test, after applying Yates's correction, to determine the statistical difference between the entire population and the sampled subset. For this test, we randomly sample the crash dataset and select 10\% of the dataset, then compare the crash count per 1-hour time intervals between the entire and the sub-sampled dataset. Examples of this count for six 1-hour intervals and three crash types are presented in this Table. The achieved p-values are higher than 0.6, meaning that there is no significant difference between the subset and the entire dataset (usually, p-values larger than 0.05 are sufficient to accept the null hypothesis). Therefore, the 5-year average of crash reports can be used for temporal analysis with a 1-day traffic video. 
To avoid sampling bias, we repeat this test for 1,000 different sub-samples and present the average results in Table \ref{tab:select_vs_entire_av}, which shows the same trend. We also performed similar tests over different weekdays, months, and years and obtained similar results suggesting that crash counts over time intervals are consistent among weekdays, months, and years. We observed the same consistency among different road segments as shown in Fig.~\ref{fig:viewcrash}.}



\begin{table}[]
\centering
\caption{\ar{Contingency Chi-Square Test with Yates's correction between the entire dataset and randomly-selected 10\% sub-sample. Each 1-hour interval is a data point.}}
\label{tab:select_vs_entire_1}
\resizebox{0.97\columnwidth}{!}{%
\blt\begin{tabular}{lllllll} \toprule
                   & \multicolumn{2}{c}{\textbf{All-type}} & \multicolumn{2}{c}{\textbf{Rear-end}} & \multicolumn{2}{c}{\textbf{Sideswipe}} \\ \midrule
Interval           & Entire            & Subset            & Entire            & Subset            & Entire             & Subset            \\ \midrule
0                  & 23                & 2                 & 7                 & 0                 & 3                  & 1                 \\
1                  & 16                & 0                 & 1                 & 0                 & 5                  & 0                 \\
2                  & 18                & 1                 & 6                 & 1                 & 2                  & 0                 \\
$\cdots$                 &   $\cdots$                  & $\cdots$                     &  $\cdots$                   &  $\cdots$                   &  $\cdots$                    &   $\cdots$                  \\
10                 & 130               & 9                 & 86                & 6                 & 36                 & 3                 \\
11                 & 168               & 16                & 92                & 8                 & 62                 & 6                 \\
12                 & 311               & 19                & 225               & 16                & 66                 & 3                 \\
$\cdots$                    & $\cdots$                    &  $\cdots$                   &$\cdots$                     & $\cdots$                    & $\cdots$                     & $\cdots$                    \\ \midrule
\textbf{Statistic} & \multicolumn{2}{c}{19.334}            & \multicolumn{2}{c}{20.359}            & \multicolumn{2}{c}{15.184}             \\
\textbf{P-value}   & \multicolumn{2}{c}{0.682}             & \multicolumn{2}{c}{0.620}             & \multicolumn{2}{c}{0.888}  \\ \bottomrule           
\end{tabular}%
}
\end{table}

\textcolor{blue}{
\begin{table}[htbp]
\centering
\caption{\ar{Contingency Chi-Square Test between the entire dataset and randomly selected 10\% subsets (as shown in Table \ref{tab:select_vs_entire_1}) after taking average over 1000 different subsets.}}
\label{tab:select_vs_entire_av}
\resizebox{0.6\columnwidth}{!}{
\blt\begin{tabular}{llll} \toprule
         & \textbf{All Type} & \textbf{Rear-end} & \textbf{Sideswipe} \\ \midrule
Statstic & 18.692& 18.073 & 18.265  \\
P-value  & 0.688& 0.716 & 0.707  \\ \bottomrule
\end{tabular}
}
\end{table}
}

\ar{To investigate the statistical difference between hours, we apply a one-way Chi-square test over different intervals averaged over the entire dataset. We repeat the test independently for all-types, rear-end, and sideswipe crashes. The P-values are extremely small for all cases suggesting that different intervals are statistically different, as noticeable in Fig. \ref{fig:viewcrash}.} 

\begin{table}[htbp]
\centering
\caption{\ar{One-way Chi-square test for 1-hour time intervals [averaged over the entire dataset], for different crash types.}}
\label{tab:chi_square_hour}
\resizebox{0.6\columnwidth}{!}{%
\blt\begin{tabular}{llll} \toprule
         & \textbf{All Type} & \textbf{Rear-end} & \textbf{Sideswipe} \\ \midrule
Statstic & 5350.39& 5028.855 & 691.545  \\
P-value  & 0.0& 0.0 & 2.679e-131  \\ \bottomrule
\end{tabular}%
}
\end{table}

\subsection{Safety metrics extraction}

The utilized trajectory extraction algorithm provides vehicle IDs, and the objects' locations (in terms of bounding box corner points per video frame), and the category of each object. We model vehicles as point objects located at the center of the bounding box. 
The extracted trajectories, after proper handling such as perspective transformation, denoising and smoothing, and filling the missing values by linear interpolation and excluding transitional and stationary non-vehicle objects, are used to compute the proposed metrics.

This information is sufficient to calculate most of the metrics including IVVR, OVVR, and OSR metrics which are solely based on the position and velocity of the vehicles. 
More particularly, the position of vehicle $n_i$ at time $t$ is $x_i(t)=\big (x_{i1}(t)+x_{i2}(t)\big )/2, y_i(t)=\big (y_{i1}(t)+y_{i2}(t)\big )/2$, where $\big (x_{i1}(t),x_{i2}(t),y_{i1}(t),y_{i2}(t)\big)$ represent the corner points of the corresponding bounding box at time point $t$. The instantaneous velocity at time $t$ is simply the derivative of the position, i.e. $v_i(t) = \sqrt{v_{ix}^2(t)+ v_{iy}^2(t)},~v_{ix}(t)=\alpha(x_i(t)-x_i(t-dt))/dt$, $v_{iy}(t)=\alpha(y_i(t)-y_i(t-dt))/dt$ with $\alpha$ being a scale factor into the desired metric unit and $dt=1/FPS$ is the time step. Higher order derivatives, and joint smoothing of the positions and velocities using methods such as Kalman filtering can also be used for higher accuracy, which we avoid in this work to keep the computational complexity at a reasonable level.

For some other metrics, such as TCI and NTC, we also need the vehicle counts and types. 
In this work, we use only two classes of vehicles: small vehicles (e.g., cars, SUVs) and large commercial vehicles (e.g., trailer trucks, busses). Since the dimensions of each object are already provided by the object detection stage (after the perspective translation and scaling), we use the object dimensions for object classification to incur minimal additional computational cost to the system. Noting the fact that the average length of personal cars and trucks is respectively about 4.5 m and 22 m \cite{car_size,truck_size}, the classification results is fairly accurate, except for the overlapping and temporarily occluded objects. For most of these objects taking the average over consecutive frames solves the transitional issues. There exist some prior work on fine-resolution vehicle classification into multiple subtypes (sedans, SUVs, truck, minivans, etc.) 
\cite{sochor2016boxcars,sochor2018boxcars,ma2019fine,ke2020fine}, which we skip here in the advantage of low computational complexity for real-time monitoring systems.
A summary of safety metrics is shown in Table \ref{tab:metrics_factor}.

\begin{table*}[]
\centering
\begin{tabular}{lll}\toprule
\multicolumn{1}{c}{\textbf{Metrics}} & \multicolumn{1}{c}{\textbf{Data Requirement}}           & \multicolumn{1}{c}{\textbf{Hyperparameter}} \\ \midrule
TTC-CV                  & Cluster Distance and Velocity        &  Cluster min distance $d_C$                      \\
IVVR                        & Velocity,\#of Vehicles               & None                         \\
OVVR                        & Velocity,\#of Vehicles               & None                         \\
OSR                         & Velocity, \#of Vehicles  & Speed Limit                         \\
TCI                         & \#of Vehicles in Each Class          & None                     \\
NTC                         & \#of Vehicles      & Vehicle's Length*                             \\
TRT                         & Timestamp                        & None                      \\ \bottomrule
\end{tabular}
\caption{The summary of proposed safety metrics along with the utilized hyperparameters. *: Vehicle's Length is calculated as 4,500mm for cars and  16,000mm for trucks.}
\label{tab:metrics_factor}
\end{table*}

\subsection{Association analysis}\label{sec:temp_analysis}

\ar{To investigate the correlation between each metric and the crash rate, we use three correlation coefficients, defined as}
\ar{
\begin{align}
\nonumber
& r_p=\frac{\sum\left(x_{i}-\bar{x}\right)\left(y_{i}-\bar{y}\right)}{\sqrt{\sum\left(x_{i}-\bar{x}\right)^{2} \sum\left(y_{i}-\bar{y}\right)^{2}}} ~~~(\text{Pearson}), \\
\nonumber
&r_s = \frac{\operatorname{cov}(\mathrm{R}(X), \mathrm{R}(Y))}{\sigma_{\mathrm{R}(X)} \sigma_{\mathrm{R}(Y)}}~~~(\text{Spearman}),\\
&\tau=\frac{2}{n(n-1)} \sum_{i<j} \operatorname{sgn}\left(x_{i}-x_{j}\right) \operatorname{sgn}\left(y_{i}-y_{j}\right) ~~~(\text{Kendall}),
\end{align}}
\ar{where $\bar{x}$, $cov(x)$, $\sigma(x)$, $R(x)$, and $\text{sgn}(x)$, are the expected value, covariance, standard deviation, rank, and sign of $x$, respectively.
Pearson's coefficient quantifies the strength of linear correlations and is most appropriate for normally distributed variables, whereas Spearman's correlation is a rank-based method that does not assume linearity or normality of the variables. Kendall offers a rank correlation based on the concordance of the pair of observations, which is less informative but more robust than the other two methods.}

\ar{We used correlation methods to find pairwise relations between the individual NSMs and crash rates. We can also use regression models to evaluate the collective prediction power of NSMs in predicting crash rate (not crash incidents).} 

\ar{It is noteworthy that crash counts can be used as approximate surrogate for risk probability, therefore the discrete values can be as the quantized [and noisy] versions of the continues-valued crash probabilities. 
This is more reasonable when crash counts are larger numbers (for 5-year crash data).} 

More specifically, we have $
y_i =\mathbf{x}_i^T\mathbf{\beta}+\varepsilon_i$, 
where $\mathbf{x}_i=[x_{i1},x_{i2},\dots,x_{iM}]$ is the set of $M$ extracted safety metrics for a given time interval, and $y_i$ is the crash count in the same interval. 
$\mathbf{\beta}=[\beta_{1},\beta_{2},\dots,\beta_{M}]$
is the vector of model parameters, and $\varepsilon_i \sim \mathcal{N}(0,\sigma^2)$ is the zero mean model noise with variance $\sigma^2$. This model is appropriate for continuous-valued outputs. To cover the count data, we also used Poisson regression model, where the crash counts $y_i$ are Poisson distributed with mean $\lambda_i$ linearly related to respective NMSs in the same interval, $x_{i1},x_{i2},\dots,x_{iM}$.    


Note that the solution of linear regression model $y_i =\mathbf{x}_i^T\mathbf{\beta}+\varepsilon_i$ (or its compact format $\mathbf{y}=X\mathbf{\beta}+\mathbf{\varepsilon}$) using the ordinary least square (OLS) is $\hat{\mathrm{\beta}}=(\mathrm{X}^T\mathrm{X})^{-1}\mathrm{X}^T.
$
From the frequentist's perspective, the true $\beta$ is deterministic but unknown. Based on these assumptions, $\hat{\beta}$ is normally distributed as \cite{faraway2004linear}:

 \begin{align*}
     \hat{\beta}&\sim N(\mathrm{X}^T\mathrm{X})^{-1}\mathrm{X}^TX\beta,(\mathrm{X}^T\mathrm{X})^{-1}\mathrm{X}^T((\mathrm{X}^T\mathrm{X})^{-1}\mathrm{X}^T)^T \varepsilon^2),\\
     \hat{\beta}&\sim N(\beta,(\mathrm{X}^T\mathrm{X})^{-1}\mathrm{X}^T\mathrm{X}(\mathrm{X}^T\mathrm{X})^{-1}\varepsilon^2),\\
     \hat{\beta}&\sim N(\beta,(\mathrm{X}^T\mathrm{X})^{-1}\varepsilon^2).
 \end{align*}
 
This justifies the validity of using the subsequent statistical analysis 
for the relevance of normally distributed model parameters. 

\begin{itemize}
\item \textbf{Adjusted R-squared}: This test is commonly used to validate the goodness of fit for linear regression models. 
R-squared statistics is defined as

 \begin{align}
     R^2 = \frac{\sum(\hat{y_i}-\Bar{y})^2}{\sum(y_i-\Bar{y})^2}
     , ~~~~0\leq R^2\leq 1,
 \end{align}
where $\hat{y}_i$ and $\Bar{y}$ are the estimated value and the average of the outputs $y_i$.  
Obviously, more predictors can result in stronger models in the presence of sufficient data samples.  
In order to account for the number of predictors and the trivial gain for using more predictors, we use Adjusted R-squared which increases when the new predictor improves the model performance more than would be expected by chance. It is defined as:

\begin{align}
    R^2_a = 1-(1-R^2)\frac{n-1}{n-p},
\end{align}
where $n$ is the number of samples and $p$ is the number of predictors.

\item \textbf{F-test}: This test evaluates the significance of the model by evaluating all observed variables simultaneously. More specifically, we have the following null hypothesis and alternative hypothesis:

    \begin{align}
        H_0:& ~\beta_1=...=\beta_k=0, \\ \nonumber
        H_a:& ~\text{Not all } \beta_j \text{ are zero}.
    \end{align}
     The F-test statistic can be computed by:
     
     \begin{align}
         F=\frac{(TSS-RSS)/(p-1)}{RSS/(n-p)}\sim F(p-1,n-p).
     \end{align}
If the achieved p-value is less than the given significance level $\alpha$, we reject the hypothesis of all model parameters being zero (irrelevant).

\item \ar{\textbf{Shapley Value:} This concept is borrowed from \textit{Game Theory} and can be used to assess the obtained value of a player (predictor) $x_i$ when combined with all other permutation of preceding players $S$ in \textit{Coalition} games \cite{afghah2018game}. In our case, NSM predictors are the players, and the value of coalition $S$ is the prediction power of a linear model constructed using these features. More specifically, we have}

\ar{\begin{align}
    \phi(x_i)=\sum_{S \subseteq \mathcal{X} \setminus {x_i}}  \frac{|\mathcal{X}-|S|-1| !|S|!}{|\mathcal{X}|!}[v(S\cup {x_i})-v(S)],
\end{align}}
\ar{where $\mathcal{X}=\{x_1,x_2,....x_M\}$ is the ordered set of NSMs, and $v(S)$ is the accuracy of linear regression model made by features $x_i \in S$. Here, $\phi(x_i)$ quantifies the added accuracy in terms of adjusted R-squared score of the model made by a set of preceding features $S$ after $x_i$ joins the coalition, $v(S\cup {x_i})-v(S)$ averaged over all computations of $S$ with preceding predictors.} 

    
    
\end{itemize}

\section{Results}



\ar{The association analysis is performed for six cameras covering six disjoint road segments as shown in Fig.\ref{fig:segment_123456}. Video is collected for 10 hours, 8:00 am - 6: 00 pm, on different days. In this analysis, each data point is a pair ($\mathbf{x}_i, y_i$) with two components: $\mathbf{x}_i$ the average of extracted safety metrics during a 10-min interval, and $y_i$ the crash count during the same interval averaged over 5 years of crash reports. Some data points are excluded due to camera issues (camera off, covered, mid-oriented).}

\begin{table}[b]
\centering
\caption{\ar{The average performance of all-predictor model relating the crash rate to all predictors (i.e. safety metrics) by 5-fold cross-validation. 
Normalized MSE (N-MSE) is calculated as $\sum \frac{(y_i-\hat{y}_i)^2}{(y_i+\hat{y}_i)^2/2}$.}} 
\label{tab:ols_analysis_1_new}
\resizebox{0.99\columnwidth}{!}{%
\blt
\begin{tabular}{lccccc}\toprule
          & \textbf{F P-value} & \textbf{$R^2$} & \textbf{adj. $R^2$} & \textbf{N-MSE (Linear)} & \textbf{N-MSE (Poisson)} \\ \midrule
All Type  & 6.52E-05           & 0.578              & 0.519                   & 0.120          & 0.122          \\
Rear-end  & 1.77E-04           & 0.565              & 0.505                   & 0.179          & 0.173          \\
Sideswipe & 2.76E-01           & 0.269              & 0.159                   & 0.536          & 0.531         \\ \bottomrule
\end{tabular}%
}

\end{table}

\subsection{\ar{Crash risk assessment}}
\begin{figure}[htbp]
\begin{center}
\centerline{\includegraphics[width=1\columnwidth]{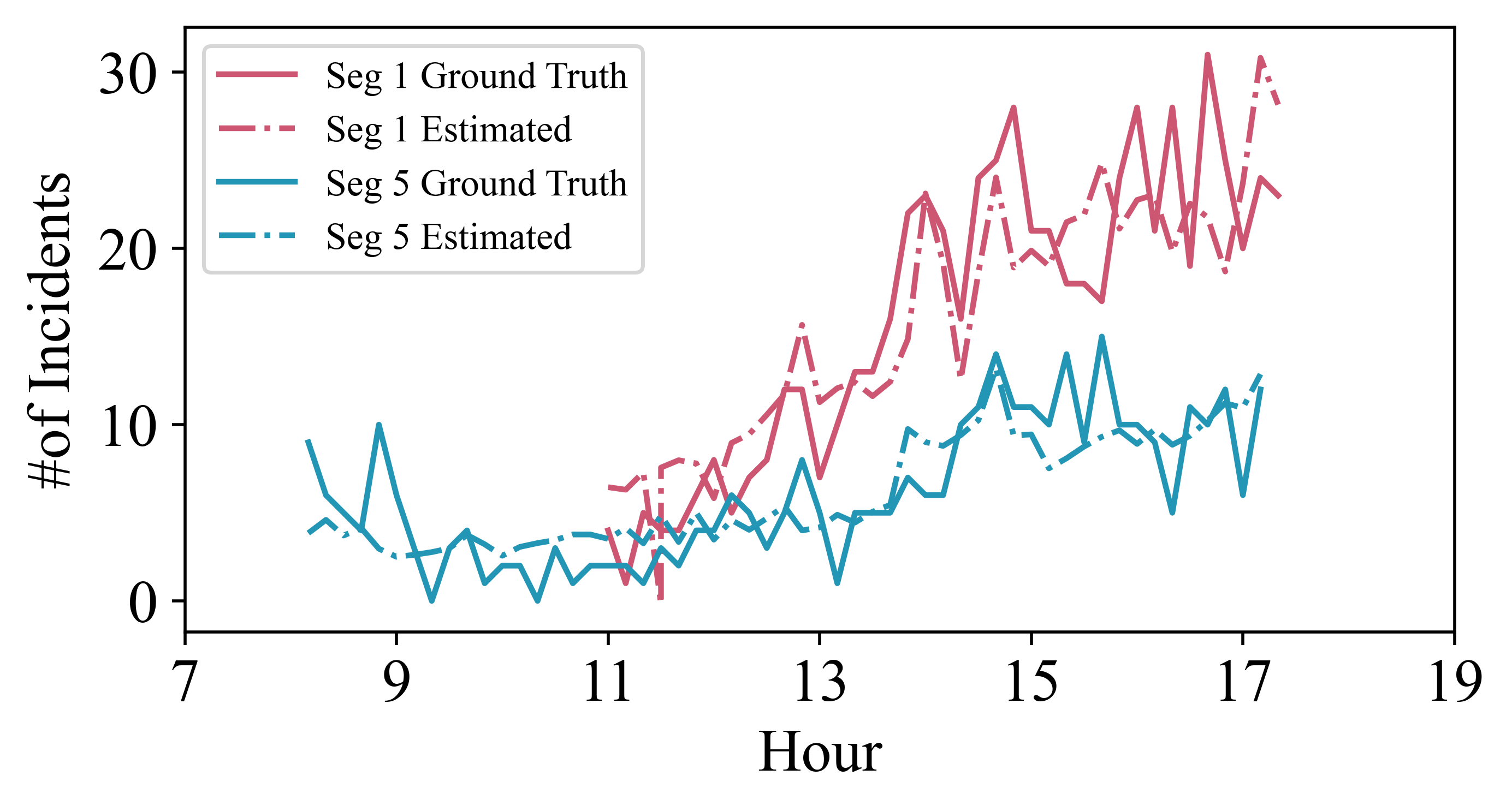}}
\caption{\ar{True and estimated rear-end crash counts for two segments using the full-predictor model.}}
\label{fig:fitted_new}
\end{center}
\end{figure}

\begin{figure}[htbp]
\begin{center}
\centerline{\includegraphics[width=0.8\columnwidth]{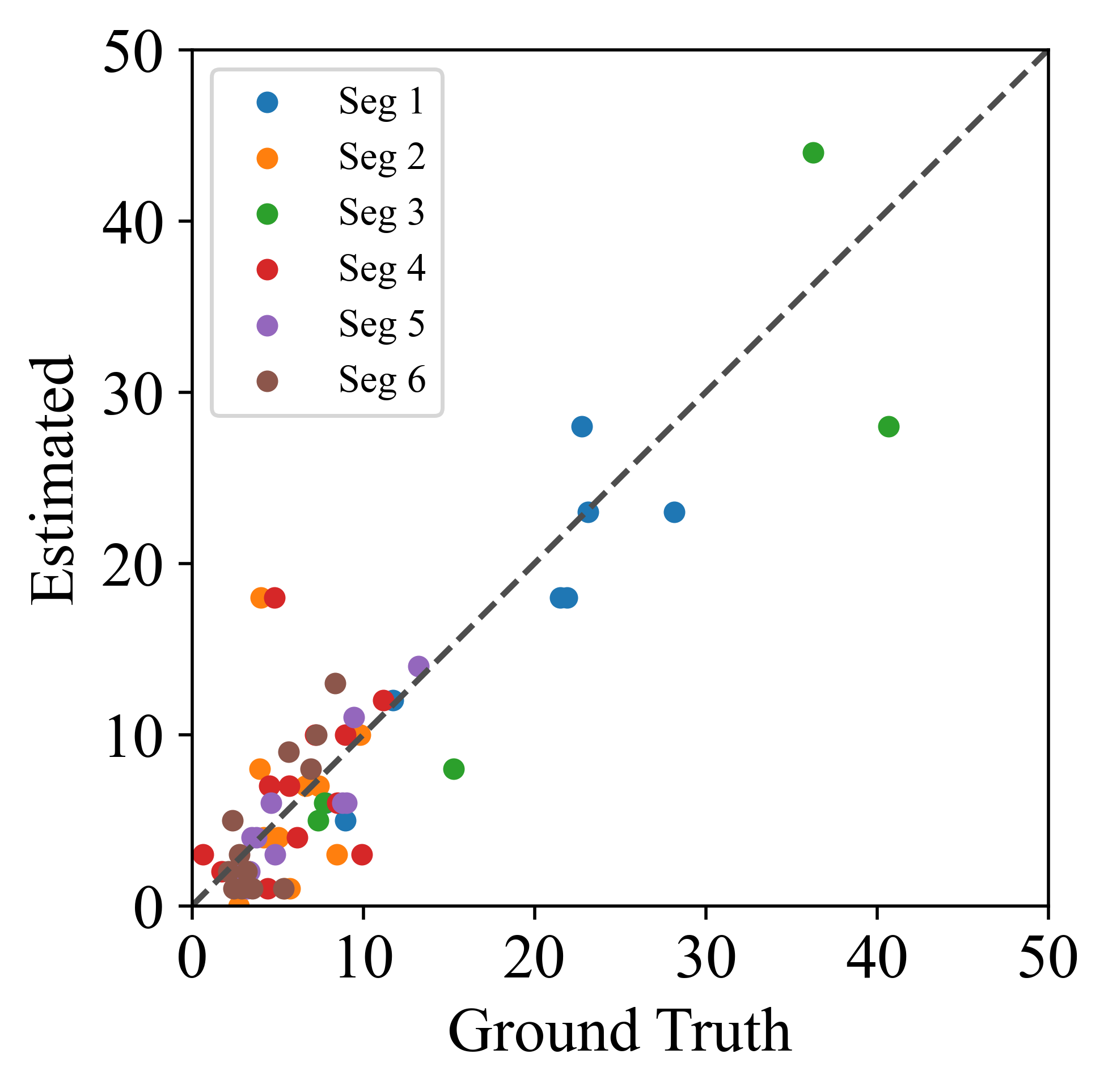}}
\caption{\ar{Estimated crash count versus ground truth for all segments. The results are for unseen samples in the validation set using the full-predictor model. Each data point represents a 10-min interval.}}
\label{fig:fitted_all_new}
\end{center}
\end{figure}

\ar{We perform this analysis to show that monitoring RSU videos and extracting NSMs can be used for risk assessment by predicting expected crash counts. To this end, we build a full-predictor model which is a linear regression model to predict the crash count based on all NSMs using 5-fold cross validation during a given interval. The prediction results are presented in Fig. \ref{fig:fitted_new}, Fig.~\ref{fig:fitted_all_new}, and Table \ref{tab:ols_analysis_1_new}.}

\begin{table*}[]
\centering
\caption{\ar{The absolute value of Pearson correlation, Spearman's correlation, and Kendall correlation for the metrics across all segments.}}
\label{tab:coeff_individual}
\resizebox{0.75\textwidth}{!}{%
\blt\begin{tabular}{clllllll|ll}\toprule
\multicolumn{1}{l}{}         &           & \multicolumn{6}{c}{\textbf{Proposed Metrics}}                                                & \multicolumn{2}{c}{\textbf{Baseline}} \\
\multicolumn{1}{l}{}         &           & \textbf{TTC-CV} & \textbf{IVVR} & \textbf{OVVR} & \textbf{OSR} & \textbf{TCI} & \textbf{NTC} & \textbf{Volume}   & \textbf{E(TTC)}   \\ \midrule
\multirow{3}{*}{Pearson}     & All Type  & 0.308           & 0.308         & 0.531         & 0.559       & 0.328       & 0.469        & 0.041            & 0.014            \\
                             & Rear-end  & 0.310           & 0.305         & 0.521         & 0.553       & 0.315       & 0.462        & 0.015            & 0.004            \\
                             & Sideswipe & 0.090           & 0.213         & 0.335         & 0.314       & 0.228       & 0.270        & 0.142            & 0.040            \\ \midrule
\multirow{3}{*}{Spearsman's} & All Type  & 0.384           & 0.453         & 0.536         & 0.538       & -0.329       & 0.512        & 0.053             & 0.014            \\
                             & Rear-end  & 0.373           & 0.429         & 0.515         & 0.535       & 0.313       & 0.503        & 0.060             & 0.004            \\
                             & Sideswipe & 0.104           & 0.330         & 0.350         & 0.283       & 0.223       & 0.247        & 0.071            & 0.040            \\  \midrule
\multirow{3}{*}{Kendall}     & All Type  & 0.273           & 0.330         & 0.377         & 0.390       & -0.246       & 0.363        & 0.045             & 0.086             \\
                             & Rear-end  & 0.265           & 0.305         & 0.358         & 0.386       & 0.235       & 0.354        & 0.050             & 0.082             \\
                             & Sideswipe & 0.091           & 0.255         & 0.267         & 0.212       & 0.165       & 0.188        & -0.044           & 0.009           \\ \bottomrule
\end{tabular}%
}
\end{table*}

\ar{Fig. \ref{fig:fitted_new} demonstrates a high alignment between the true and predicted rear-end crash counts using the full-predictor model. The same results are shown for all-type crash counts for all six segments as a scatter plot in Fig. \ref{fig:fitted_all_new}, where most of the data samples concentrate around the unit-slope line (true value=predicted value). The results verify the usefulness of the proposed approach of using safety metrics to predict crash rates.}

A more formal statistical analysis is provided in Table \ref{tab:ols_analysis_1_new} using the metrics discussed in section \ref{sec:nms} to validate the relevance of the constructed linear regression model. The results are based on the average of the 5-fold cross-validation \ar{(instead of the best results)}, which further confirms the validity of the developed models. \ar{It can be seen that} the P-value is less than 0.05 for all crash types (using three different tests), which shows that the combination of all predictors (NSMs) makes a non-zero (significant) contribution to predicting crash rates, with any reasonable significance value.
\ar{Similarly, the normalized MSE for both linear regression model as well as Poisson regression model are in an acceptable range for all-type and rear-end crashes. The results suggest that predicting crashes based on NSMs is more relevant for rear-end crashes than the sideswipes. This is justifiable since most metrics (such as TTC, TTC-CV, IVVR, OVVR, OSR) consider longitudinal motions while sideswipe crashes highly depend on latitudinal motions. 
This reveals the need for developing a richer set of network-level safety metrics that are capable of predicting sideswipe metrics. such as metrics that indicate zigzag driving can be helpful. 
Also, sideswipe crashes seem to be more complicated in nature and potentially depend on other factors such as human mistakes.}


\subsection{Temporal correlation between NMS and crash count}

\begin{figure}[]
\begin{center}
\centerline{\includegraphics[width=1\columnwidth]{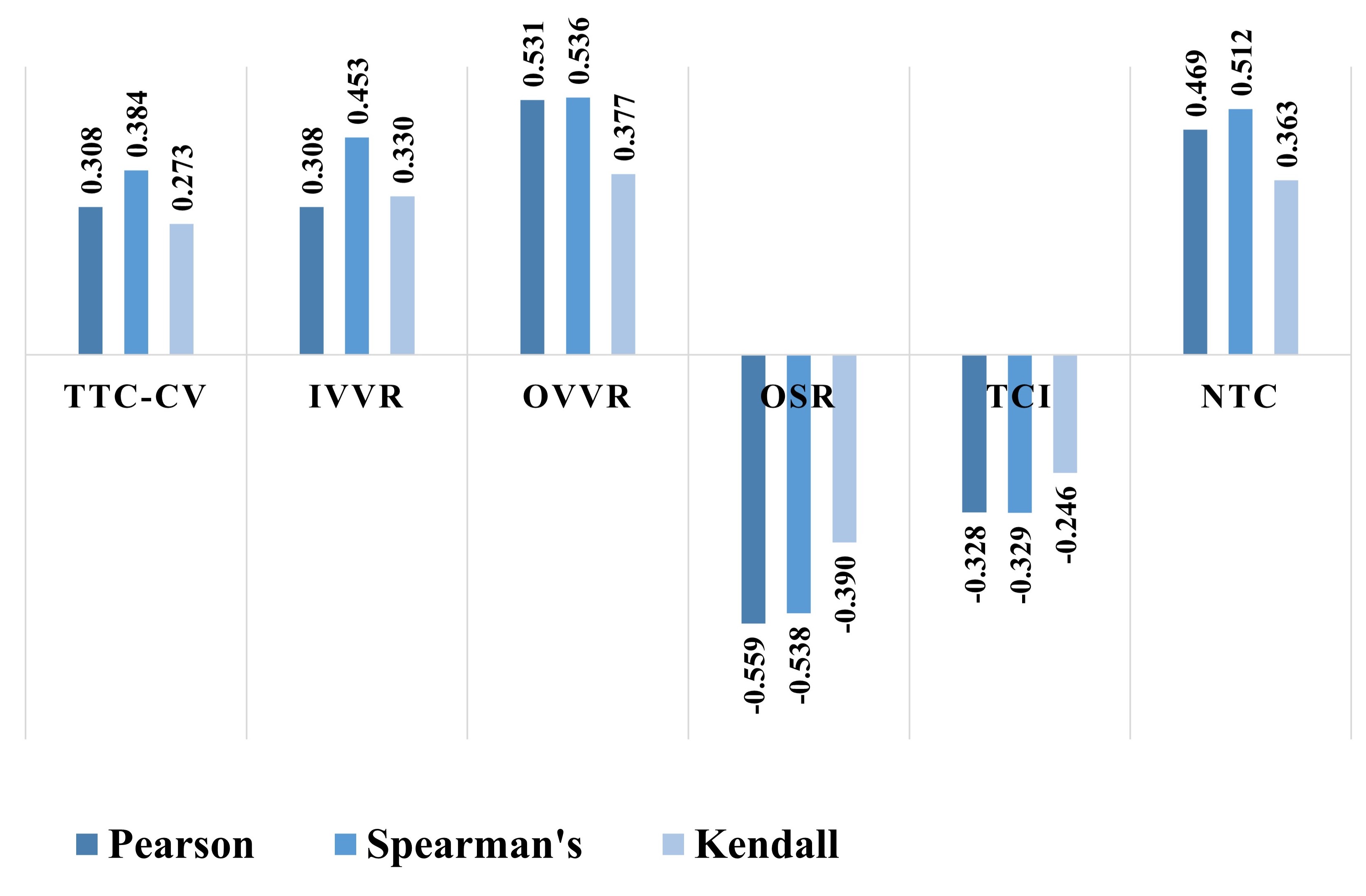}}
\caption{\ar{Individual correlations between different metrics and all-types crash counts.}}
\label{fig:single_all_type}
\end{center}
\end{figure}

\ar{Now that the collective power of NSMs in predicting crash rates is established, we develop further analysis to estimate the relevance of each metric individually.}
To this end, we compute the individual correlation of each safety metric during a given interval with any of the crash types for each segment during the same time interval, using Pearson, Spearman, and Kendall correlation measures. \ar{To further highlight the importance of the achieved correlations, we compare the results against two reference values, (i) the correlation between the crash types and the traffic volume, and (ii) the correlation between the crash count and individual TTCs averaged over the entire traffic, E(TTC) discussed in Section \ref{sec:intro}. The result are presented in Table \ref{tab:coeff_individual} and Fig. \ref{fig:single_all_type}, showing that the proposed metrics exhibit a high correlation ($0.25\sim0.5$) for rear-end and all-type crashes, which is much higher than the baseline correlation between the traffic volume and crash counts ($0.01 \sim 0.08$). Also, it is clearly seen that the achieved correlation for TTC-CV is in range ($0.09\sim 0.38$), much higher than that of E(TTC) in range ($0.004 \sim 0.08$). This observation supports the idea of developing new network-level safety metrics.}

\begin{table*}[]
\centering
\caption{\ar{The average absolute Pearson correlation between NSMs and crash count across segments. The numbers in parentheses denote the number of combinations.}}
\label{tab:cross-segment}
\resizebox{0.5\textwidth}{!}{%
\blt\begin{tabular}{clrrrrrr}\toprule
\multirow{2}{*}{\textbf{Scenario}} & \multicolumn{1}{c}{\multirow{2}{*}{\textbf{Type}}} & \multicolumn{6}{c}{\textbf{Metrics}} \\
 & \multicolumn{1}{c}{} & \multicolumn{1}{l}{\textbf{TTC-CV}} & \multicolumn{1}{l}{\textbf{IVVR}} & \multicolumn{1}{l}{\textbf{OVVR}} & \multicolumn{1}{l}{\textbf{OSR}} & \multicolumn{1}{l}{\textbf{TCI}} & \multicolumn{1}{l}{\textbf{NTC}} \\ \midrule
\multirow{3}{*}{\begin{tabular}[c]{@{}c@{}}Individual \\      segments\\      (\#6)\end{tabular}} & All-Type & 0.308 & 0.308 & 0.531 & 0.599 & 0.328 & 0.469 \\
 & Rear-end & 0.310 & 0.307 & 0.521 & 0.591 & 0.315 & 0.462 \\
 & Sideswipe & 0.090 & 0.211 & 0.335 & 0.329 & 0.228 & 0.270 \\ \midrule
\multirow{3}{*}{\begin{tabular}[c]{@{}c@{}}Combine   2\\       segments\\      (\#15)\end{tabular}} & All-Type & 0.297 & 0.380 & 0.550 & 0.575 & 0.441 & 0.412 \\
 & Rear-end & 0.286 & 0.384 & 0.536 & 0.568 & 0.411 & 0.399 \\
 & Sideswipe & 0.176 & 0.281 & 0.430 & 0.403 & 0.380 & 0.304 \\ \midrule
\multirow{3}{*}{\begin{tabular}[c]{@{}c@{}}Combine   3\\       segments\\      (\#20)\end{tabular}} & All-Type & 0.268 & 0.377 & 0.536 & 0.553 & 0.435 & 0.350 \\
 & Rear-end & 0.253 & 0.387 & 0.522 & 0.546 & 0.404 & 0.336 \\
 & Sideswipe & 0.186 & 0.276 & 0.445 & 0.415 & 0.403 & 0.291 \\ \midrule
\multirow{3}{*}{\begin{tabular}[c]{@{}c@{}}Combine   4\\       segments\\      (\#15)\end{tabular}} & All-Type & 0.238 & 0.384 & 0.529 & 0.540 & 0.421 & 0.309 \\
 & Rear-end & 0.222 & 0.395 & 0.515 & 0.534 & 0.389 & 0.294 \\
 & Sideswipe & 0.184 & 0.276 & 0.451 & 0.420 & 0.407 & 0.279 \\ \midrule
\multirow{3}{*}{\begin{tabular}[c]{@{}c@{}}Combine   5\\       segments\\      (\#6)\end{tabular}} & All-Type & 0.212 & 0.396 & 0.526 & 0.534 & 0.408 & 0.281 \\
 & Rear-end & 0.196 & 0.408 & 0.513 & 0.527 & 0.376 & 0.265 \\
 & Sideswipe & 0.181 & 0.280 & 0.455 & 0.422 & 0.409 & 0.270 \\ \midrule
\multirow{3}{*}{\begin{tabular}[c]{@{}c@{}}Combine   all\\       segments\\      (\#1)\end{tabular}} & All-Type & 0.191 & 0.412 & 0.526 & 0.530 & 0.397 & 0.262 \\
 & Rear-end & 0.174 & 0.424 & 0.513 & 0.524 & 0.366 & 0.246 \\
 & Sideswipe & 0.178 & 0.288 & 0.458 & 0.424 & 0.409 & 0.265\\ \bottomrule
\end{tabular}%
}
\end{table*}

\begin{table}[]
\centering
\caption{\ar{The cross-segment analysis for full model. The model is trained using 5 segments and tested with the unseen segment}.}
\label{tab:cross_analysis_1_new}
\resizebox{0.9\columnwidth}{!}{%
\blt\begin{tabular}{lcccc}\toprule
          & \textbf{F P-value} & \textbf{$R^2$} & \textbf{adj. $R^2$} & \textbf{N-MSE} \\ \midrule
All Type  & 7.20E-31           & 0.510              & 0.499                   & 0.219            \\
Rear-end  & 1.35E-28           & 0.507              & 0.496                   & 0.287            \\
Sideswipe & 3.51E-14           & 0.295              & 0.278                   & 0.622           \\ \bottomrule
\end{tabular}%
}
\end{table}

\subsection{Spatial analysis}
\ar{So far, we showed the high correlation between the NSMs and crash rates of different types, by investigating each segment individually. To investigate the generalization of the proposed method, we perform cross-segment analysis. Specifically, we perform the same correlation analysis to all combinations of 2 to 5 segments, as shown in Table \ref{tab:cross-segment}. the results are consistent with Table \ref{tab:coeff_individual} that shows a reasonable consistency across segments.} 

\ar{To further investigate that the spatial generalizability of the results, we perform a cross-segment analysis. We construct a linear model (to predict crash counts based on NSMs) for five segments, and evaluate the model for the remaining segment (out-of-segment validation). Then, we repeat the test for all other combinations, so that each segment is tested once. The results are shown in Table \ref{tab:cross_analysis_1_new}.}

\subsection{Coalitions of predictors}
\ar{It is known that the predictive power of each feature can depend on the presence of other features, due to inter-feature linear and non-linear correlations \cite{afghah2018game}. To account for this fact, we calculate the Shapley value for each metric as discussed in section \ref{sec:temp_analysis}. Here, $v(S)$ is the value of a coalition $S$ calculated as the adjusted R-squared statistics of a model built using predictors $x_i \in S$. The results are presented in Table \ref{tab:Shapely_value} that show these metrics play a relatively balanced role in the cooperative operation. Therefore, it is advantageous to use the full-predictor model based on all proposed NSMs.}

\begin{table}[]
\centering
\caption{\ar{The Shapley value of the proposed metrics for different types of crashes. }}
\label{tab:Shapely_value}
\resizebox{\columnwidth}{!}{%
\blt\begin{tabular}{lllllll} \toprule
          & \textbf{TTC-CV}    & \textbf{IVVR}      & \textbf{OVVR}      & \textbf{OSR}       & \textbf{TCI}       & \textbf{NTC}       \\ \midrule
All Type  & 2.493E-03 & 1.975E-03 & 2.557E-03 & 4.886E-03 & 2.744E-03 & 2.576E-03 \\
Rear-end  & 2.450E-03 & 1.956E-03 & 2.382E-03 & 4.892E-03 & 2.557E-03 & 2.557E-03 \\
Sideswipe & 2.263E-04 & 6.377E-04 & 1.318E-03 & 1.236E-03 & 1.029E-03 & 5.294E-04 \\ \bottomrule
\end{tabular}%
}
\end{table}

Overall, we made the following observations, some of which require further investigations using more data samples to draw stronger conclusions. i) Monitoring RSU traffic video and extracting network-level safety metrics can be used for crash risk analysis; ii) The models build for dome road segments are generalizable to roads with similar geometry; iii) OSR exhibits a negative correlation with crash count, which is counter intuitive and can reflect the fact that crashes are more likely in busy hours than light traffics. This requires further investigation with larger datasets; iv) traffic composition represented by TCI shows that a more unbalanced traffic flow (extremely different number of small cars and trucks) is more prone to making crashes; 
v) cluster level analysis of TTC presents credibility in analyzing the traffic safety since it participates in most of the restricted models, vi) sideswipe crashes are harder to predict with metrics driven from longitudinal motions, and perhaps human factors or road geometry play more significant roles in modulating crash rates.



\section{Conclusion}
In this paper, we offered a set of network-level safety metrics to assess the overall safety characteristics of traffic flow in a given driving zone. This concept extends the popular notion of safety metrics to network-level analysis. We showed that the proposed safety metrics are highly correlated with crash frequency (temporally and spatially). 
We conducted a case study in the state of Arizona by integrative analysis of collected video files from the I-10 highway and 5-year crash reports that verify the association between the network-level safety metrics and crash frequency in the same time intervals. More specifically, metrics that gauge the overall speed variation of vehicles, the traffic composition and diversity of vehicles, the density of traffic volume, and also relative mobility of car clusters are highly correlated with the crash frequency (with p-values much lower than 5\% for most scenarios). We also observed that rear-end crashes are easier to predict than side-swipe crashes, perhaps due to the stronger role of road geometry and human mistakes in side-swipe accidents. Also, it shows the need for developing safety metrics that mimic latitudinal motions in addition to longitudinal motions. 

\ar{The practical use of this analysis is identifying risk factors by constant monitoring of traffic flow using AI-based roadside infrastructures to broadcast warning messages and take more efficient traffic control decisions. Also, traffic control teams can take redesign and long-term decisions to keep the safety metrics in an acceptable range to enhance the overall driving safety on the road. 
Developing lightweight deep learning models to process traffic video and extract safety metrics in a real-time fashion can pave the road for developing online risk assessment systems.}

\section{Declaration of Competing Interest}
The authors declare that they have no known competing financial interests or personal relationships that could have appeared to influence the work reported in this paper.

\section{Acknowledgment}
We would like to thank the NSF and the Institute of Automated Mobility (IAM) for supporting this work. We also thank the Arizona Department of Transportation (ADOT) for sharing roadside infrastructure, crash reports, and other resources with us during the performance of this project. Also, the opinions, findings, and conclusions expressed in this manuscript are those of the author’s and not necessarily those of the IAM and ADOT.

 \bibliographystyle{elsarticle-num} 
 \bibliography{cas-refs}
 
\ifx \includeAppendix \incYes
\appendix

\section{Taxonomy for Operational Safety Metrics } \label{sec:safety-metrics}
An essential objective of operational  safety analysis from a scenario perspective, as opposed to the network-level perspective that is the focus of this paper, is to implement \textit{operational safety assessment (OSA) metrics}, which are quantifiable measures extracted from traffic videos (or other data sources). These OSA metrics allow for a determination of the operational safety of a vehicle (AV or human-driven) to be assessed as a given scenario is navigated. Here, we review key OSA metrics that have been broadly used for operational safety analysis. It is notable that many algorithms developed for AVs utilize safety metrics for safe navigation decisions and avoiding crashes; however, we consider both self-driving and human-driven vehicles.

In this Appendix, for the sake of completeness, we review key OSA metrics that have been broadly used for operational safety analysis along with the level of access to ADS data to extract these metrics for AVs. 
A recent paper \cite{mahmud2017application} summarizes safety metrics according to different basis (temporal, distance, and deceleration):
\begin{enumerate}[i]
  \item \textbf{Temporal-based indicators:} Time to Collision (TTC), Extended Time to Collision (Time Exposed Time-to-Collision(TET), Time Integrated Time-to-Collision(TIT)\cite{minderhoud2001extended}), Modified TTC (MTTC), Crash Index (CI), Time-to-Accident (TA), Time Headway (THW), and Post-Encroachment Time (PET). 
  \item \textbf{Distance-based indicators:} Potential Index for Collision with Urgent Deceleration (PICUD), Proportion of stopping Distance (PSD), Margin to Collision (MTC), Difference of Space Distance and Stopping Distance (DSS), Time Integrated DSS (TIDSS), and Unsafe Density (UD);
  \item \textbf{Deceleration-based indicators:} Deceleration Rate to Avoid a Crash (DRAC), Crash Potential Index(CPI), and Criticality Index Function (CIF).
\end{enumerate}

A more recent paper \cite{wishart2020driving} by the  metric team of the Institute of Automated Mobility (IAM), 
provides a comprehensive set of OSA metrics following an extensive literature review. The objective was to develop a set of metrics for both human-driven and AVs that includes existing, adapted, and novel metrics. In a follow-up paper\cite{elli2021evaluation}, the IAM proposed a taxonomy for operational safety metrics that is explored and expanded upon here. The IAM work is also a component of an Recommended Practice standards being developed by the SAE Verification and Validation (V\&V) Task Force under the On-Road Automated Driving (ORAD) Committee\cite{OSM2020}. The taxonomy will be introduced first, and then selected OSA metrics will be discussed (including the metrics from \cite{mahmud2017application} listed above).

\begin{figure}[]
\begin{center}
\centerline{\includegraphics[width=1\columnwidth]{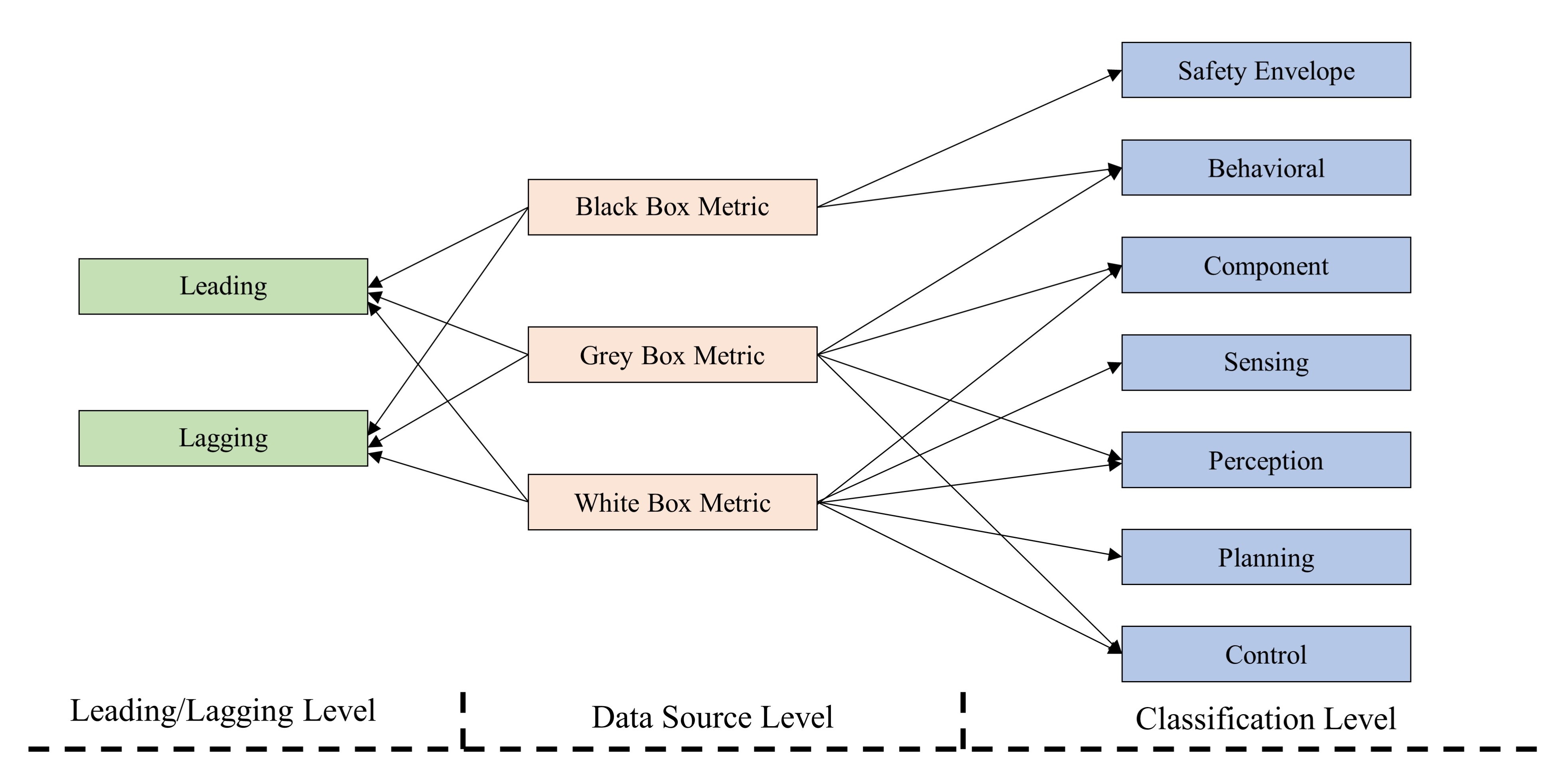}}
\caption{Operational safety metrics taxonomy proposed by the IAM.}
\label{fig:taxonomy}
\end{center}
\end{figure}

The expanded taxonomy introduced here is shown in Figure.\ref{fig:taxonomy}. The highest taxonomic rank in the proposed taxonomy hierarchy consists of three types that are based, essentially, on the data source, which includes the level of access required of ADS data. This access to proprietary data could be challenging, depending on the implementation of the OSA metric; it should be noted that “lighter” metrics require more cooperation with the AV developer. The three types are (example metrics are given for each, and are described in more detail later in the section):
\begin{enumerate}
\item \textbf{Black Box Metric:} A metric that allows measurement of data that can be obtained without requiring any access to ADS data. This could be from an on-board or off-board source. ADS data may enhance the accuracy and precision of the measurement(s).
EXAMPLE:Collision incident (CI).

\item \textbf{Grey Box Metric:} A metric that allows measurement of data that can only be obtained with limited access to ADS data.
EXAMPLE: ADS DDT Execution (ADE).

\item \textbf{White Box Metric:} A metric that allows measurement of data that can only be obtained with significant access to ADS data.
EXAMPLE: Perception Precision (PP).
\end{enumerate}

The second rank in the taxonomic hierarchy is the classification rank, which consists of the following (again, an example metric that is described later in the section is included):
\begin{enumerate}
\item\textbf{Safety Envelope Metric:} A metric that allows for measurement of the subject vehicle’s maintenance of a safe boundary around itself. This includes situations that may not be within the subject vehicle’s control.
		EXAMPLE: Minimum Safety Envelope (MSE).
\item\textbf{Behavioral Metric:} A metric that allows for measurement of an improper behavior of the subject vehicle.
EXAMPLE: Aggressive Driving (AD).
	
\item\textbf{Component Metric:} A metric that allows for measurement of the proper function of ADS components.
EXAMPLE: Event Data Recorder Compliance (EDRC).
	
\item\textbf{Sensing Metric:} A metric that allows for measurement of the ability of the ADS sensors to receive adequate information from the AV environment.
EXAMPLE: Camera Resolution (CR).
	
\item\textbf{Perception Metric:} A metric that allows for measurement of the ability of the ADS to interpret information about the AV environment obtained by the ADS sensors.
EXAMPLE: Human Traffic Control Detection Error Rate (HTCDER).
	
\item\textbf{Planning Metric:} A metric that allows for measurement of the ability of the ADS to plan an appropriate route through the AV environment.
EXAMPLE: Object Avoidance Plan Error (OAPE).
	
\item\textbf{Control Metric:} A metric that allows for measurement of the ability of the AV to execute the planned route devised by the ADS.
EXAMPLE: Actuation Error (AE).

\end{enumerate}

It is important to note that not all classification rank types link to data source rank types in the current version. For example, the Component Metric is not linked to the Black Box Metric because data at the component level is deemed to be more access to (potentially proprietary) data than is allowed for the Black Box Metric.

The third rank in the taxonomy hierarchy is the Leading/Lagging rank, which relates (in binary fashion) to either the prediction (i.e., before) of a potential operational safety outcome or report (i.e., after) of an operational safety outcome after it has occurred. Operational safety outcomes include conflicts, collisions, an ADS disengagement, or a violation of a traffic law:
\begin{enumerate}
\item \textbf{Leading:} An OSA identification of a potential future operational safety outcome. 
EXAMPLE: Any safety envelope metric.

\item \textbf{Lagging:} An OSA identification of a report of an operational safety outcome.
EXAMPLE: Any traffic law violation metric.

\end{enumerate}

The metrics currently being considered for inclusion in the SAE J3237 Recommended Practice\footnote{This SAE standards document is currently in development and the operational safety metrics list could be modified, expanded, or contracted. 
It should be noted that the objective for the SAE J3237 document is to identify the minimum number of safety envelope metrics that will allow for an unsafe situation to be identified. Ideally, a single safety envelope metric would suffice; however, two or more safety envelope metrics may be necessary. The IAM is currently conducting research into this question of which safety envelope metrics to select, and the result may be adopted in the SAE J3237 Recommended Practice document.} are included in Table \ref{tab:taxonomy} (although the list may change)). The IAM has focused on Black Box Metrics and Grey Box Metrics as part of the comprehensive set introduced in \cite{mahmud2017application}.\\
\begin{table*}[]
\resizebox{0.8\textwidth}{!}{
\centering%
\begin{tabular}{lcc}\toprule
\multicolumn{1}{c}{\textbf{Metric Name}}               & \textbf{\begin{tabular}[c]{@{}c@{}}Data \\ Source \\ Taxonomy\end{tabular}} & \textbf{\begin{tabular}[c]{@{}c@{}}Classification\\  Taxonomy\end{tabular}} \\ \midrule
Minimum Safe Envelope (MSE)                   & Black                                                                       & Safety Envelope                                                             \\
Proper Response  (PR)                             & Black                                                                       & Safety Envelope                                                             \\
Minimum Safe Distance Factor (MSDF)                      & Black                                                                       & Safety Envelope                                                             \\
Time-to-Collision (TTC)                                  & Black                                                                       & Safety Envelope                                                             \\
Modified Time-to-Collision (MTTC)                        & Black                                                                       & Safety Envelope                                                             \\
Time Headway (TH)                                        & Black                                                                       & Safety Envelope                                                             \\
Instantaneous Safety Metric (ISM)                        & Black                                                                       & Safety Envelope                                                             \\
Collision Avoidance Capability (CAC)                     & Black                                                                       & Safety Envelope                                                             \\
Minimum Safe Distance Infringement (MSDI)                & Black                                                                       & Safety Envelope                                                             \\
Post-Encroachment Time (PET)                             & Black                                                                       & Safety Envelope                                                             \\
Not-at-Fault Collision Incident (NAFCI)                  & Black                                                                       & Safety Envelope                                                             \\
Rear-End Collision Incident (RECI)                       & Black                                                                       & Safety Envelope                                                             \\
Deceleration Rate to Avoid the Crash (DRAC)              & Black                                                                       & Safety Envelope                                                             \\
Difference of Space Distance and Stopping Distance (DSS) & Black                                                                       & Safety Envelope                                                             \\
Maximum Speed (MaxS)                                     & Black                                                                       & Behavioral                                                                  \\
Delta Velocity ($\Delta$V)                                      & Black                                                                       & Behavioral                                                                  \\
Traffic Law Violation (TLV)                        & Black                                                                       & Behavioral                                                                  \\
Evasive Action (EA)                                      & Black                                                                       & Behavioral                                                                  \\
Aggressive Driving (AD)                                  & Black                                                                       & Behavioral                                                                  \\
Collision Incident (CI)                                  & Black                                                                       & Behavioral                                                                  \\ \midrule
ADS DDT Execution (ADE)                                  & Grey                                                                        & Control                                                                     \\
Achieved Behavioral Competency (ABC)                     & Grey                                                                        & Planning                                                                  \\
Human Traffic Control Perception Error Rate (HTCDER)     & Grey                                                                        & Perception                                                                  \\
Human Traffic Control Violation Rate (HTCVR)             & Grey                                                                        & Behavioral                                                                  \\
Minimum Safe Distance Calculation Error (MSDCE)          & Grey                                                                        & Perception                                                                  \\
Safety-Related Component Failure Perception (SRCFD)      & Grey                                                                        & Component                                                                   \\
Event Data Recorder Compliance (EDRC)                    & Grey                                                                        & Component                                                                   \\ \midrule
Perception Precision (PP)                                & White                                                                       & Perception                                                                  \\
Perception  Rate (PDR)                         & White                                                                       & Perception                                                                  \\
Perception Weighted Harmonic Mean (PWHM)                 & White                                                                       & Perception                                                                  \\
Perception False Positive Rate (PFPR)                    & White                                                                       & Perception                                                                  \\
Perception False Negative Rate (PFNR)                    & White                                                                       & Perception                                                                  \\
Anomaly Perception Behavior (ADB)                        & White                                                                       & Perception                                                                  \\
Multiple Object Perception Error Rate (MODER)            & White                                                                       & Perception                                                                  \\
Localization Error (LE)                                  & White                                                                       & Perception                                                                  \\
Multiple Object Tracking Precision (MOTP)                & White                                                                       & Perception                                                                  \\
Data Conflict Perception Rate (DCDR)                     & White                                                                       & Perception                                                                  \\
Intersection over Union (IoU)                            & White                                                                       & Perception                                                                  \\
Actuation Error (AE)                                     & White                                                                       & Control                                                                   \\
Software Execution Error (SEE)                           & White                                                                       & Component                                                                   \\
Object Avoidance Plan Error (OAPE)                       & White                                                                       & Planning       \\ \bottomrule 
\end{tabular}%
}
\caption{Current SAE J3237 Information Report operational safety metrics.}
\label{tab:taxonomy}
\end{table*}
\indent We also mention the approach an infrastructure-based (i.e., off-board the vehicle) observer system takes to monitor video and extract OSA metrics. The observer system also called system in the rest of this appendix for convenience, is a terrestrial or aerial monitoring system that collects traffic video for processing from an external observer’s point of view. A list of these metrics, along with their brief descriptions, is presented in Table \ref{tab:metrics}. 

Our main references include \cite{wishart2020driving}, \cite{mahmud2017application}, and \cite{gettman2003surrogate}.

\begin{table*}[]
\resizebox{\textwidth}{!}{%
\begin{tabular}{lll}\toprule
\textbf{Metric}     & \textbf{Definition} & \textbf{Features}                                                                                                                         \\ \midrule
MaxS*       & $\max v$         & A simple but effective measure of the severity of a collision.                                                                   \\ \midrule
$\Delta v$*    & $\Delta v$          & An effective indicator of the severity of a collision.                                                                           \\ \midrule
TTC &
  $TTC =\frac{X_L-X_F}{v_F-v_L}$ &
  \begin{tabular}[c]{@{}l@{}}Provides more information than PET; assumes that involved vehicles are at a constant speed;\\ cannot reflect the severity.\end{tabular} \\ \midrule
PET &
  $PET =t_2-t_1$ &
  \begin{tabular}[c]{@{}l@{}}Can be easily captured and computed; suitable for assessing intersecting conflicts; \\ only suitable for the cases of transversal trajectories;\\ cannot reflect the severity.\end{tabular} \\ \midrule
Initial DR & $DR_{t=0}$          & A basic indicator; cannot consider the following vehicle's acceleration/deceleration.                                                        \\ \midrule
DRAC       &$TTC =\frac{(v_F-v_L)^2}{X_L-X_F}$         & \begin{tabular}[c]{@{}l@{}}No need to directly observe the deceleration rate; \\ does not work for lateral movement\end{tabular}      \\ \midrule
CPI*       & $CPI_i=\frac{\sum_{t=t_i}^{tf_i}P(DRAC_{i,t})\geq MADR\Delta t\cdot b}{T_i}$          & Considers more factors such as traffic and road conditions compared to DRAC;                                                         \\ \midrule
PSD*        &\begin{tabular}[c]{@{}l@{}}$PSD =\frac{RD}{MSD}$;\\ $ PSD =\frac{v^2}{2MADR} $ \end{tabular}      & \begin{tabular}[c]{@{}l@{}}Works for a conflict with a single vehicle involved; \\ rarely used for specific safety problems.\end{tabular} \\ \midrule
UD &
  $UD=\frac{\sum_{S=1}^{S_t}\sum_{V=1}^{V_t}unsafety_{V,S}\cdot d}{TL}$ &
  \begin{tabular}[c]{@{}l@{}} Delivers more accurate information than typical micro-simulation outputs;\\ allows comparison between simulation scenarios or links;\\ some parameters are difficult to capture automatically;\\ only applicable to identical trajectories.\end{tabular} \\ \midrule
  
  MSE &
  \begin{tabular}[c]{@{}l@{}}Indicate the subject vehicle violate safety boundary of another.\\ Safety boundary: $d_{min}^{long,same}$(Eq.\ref{eq:d_long_same}) $d_{min}^{long,opp}$(Eq.\ref{eq:d_long_opp}) and $d_{min}^{lat}$(Eq.\ref{eq:d_lat})\end{tabular} &
  \begin{tabular}[c]{@{}l@{}}Provides how to calculate the quantified risky distances;\\ a basis of many assessment methods.\end{tabular} \\ \midrule
PR&
  \begin{tabular}[c]{@{}l@{}}An action to recover when\\ $d_{min}^{long},d_{min}^{lat}$ and the safe range of $a^{lat}$,$a^{long}$ are violated\end{tabular} &
  Adds more information beyond MSDV. \\ \midrule
MSDF &
  \begin{tabular}[c]{@{}l@{}}$MSDF^{lat} = \frac{d^{lat}}{d^{lat}_{min}}$\\ $MSDF^{long} = \frac{d^{long}}{d^{long}_{min}}$\end{tabular} &
  \begin{tabular}[c]{@{}l@{}}Can indicate the degree of defensive driving style;\\ a higher $MSDF$ may not means higher safety.\end{tabular} \\ \midrule
MSDCE &
  $MSDCE = \sqrt{\frac{\vert d^{long}_{gt,min} - d^{long}_{min}\vert }{d^{long}_{gt,min}}+\frac{\vert d^{lat}_{gt,min} - d^{lat}_{min}\vert }{d^{lat}_{gt,min}}}$ &
  Indicate the ADS ability to determine the safety distances. \\ \midrule
CI &
  \begin{tabular}[c]{@{}l@{}}Indicate the subject vehicle is involved in a collision.\\ Active when $d^{lat}$ and $d^{long}$ are equal to 0.\end{tabular} &
  \begin{tabular}[c]{@{}l@{}}Capture instances of collisions as a metric;\\ the severity of the collision can be defined by KABCO scale\cite{herbel2010highway}:\\
  K(Fatal Injury), A(Incapacitating Injury), B (Non-Incapacitating Injury),\\ C(Possible Injury), O(No Injury).\end{tabular} \\ \midrule
TLV &
  Indicate the subject vehicle that violates a traffic law &
  \begin{tabular}[c]{@{}l@{}}Emphasizes that an ADS vehicle must follow existing laws;\\ exceptions are made, for example, when road closures \\ require temporary violations of driving exclusively inside a traffic lane.\end{tabular} \\ \midrule
ABC &
  Indicate the subject vehicle can execute a specific behavior correctly. &
  \begin{tabular}[c]{@{}l@{}}Indicate the safety of ADS;\\ included in the preliminary list of metrics.\end{tabular} \\ \midrule
ADSA &
  Indicate the ADS is active when executing behaviors. &
  \begin{tabular}[c]{@{}l@{}}Indicate the safety of ADS;\\ included in the preliminary list of metrics;has some dispute in California\cite{hawkins2019california}.\end{tabular} \\ \midrule
HTCDER &
  $HTCDER = \frac{GTI-CDI}{CDI}$ &
  \begin{tabular}[c]{@{}l@{}}One of HTC measurements;\\ manually instructions by officers should still be considered.\end{tabular} \\ \midrule
HTCVR &
  $HTCVR = \frac{CDI-CCI}{CDI}$ &
  \begin{tabular}[c]{@{}l@{}}One of HTC measurements;\\ manually instructions by officers should still be considered.\end{tabular} \\ \midrule
  
AD &
  \begin{tabular}[c]{@{}l@{}}Indicate the maneuvers (longitudinal/lateral accelerations) \\ of a subject vehicle exceeds specified thresholds.\end{tabular} &
  \begin{tabular}[c]{@{}l@{}}The thresholds vary by jurisdiction and culture;\\ AD is only involved the subject vehicle;\\ implies the inherent and potential risks of natural driving behaviors;\\ should be included when evaluating ADS.\end{tabular} \\ \bottomrule \end{tabular}}
\caption{Summary of operational safety metrics along with their key properties. Noting that '*' denotes the metrics that are not selected by IAM but are basic metrics employed in other papers.}  
\label{tab:metrics}
\end{table*}


\subsection{Summary of Basic Operational Safety Metrics}

~~~\textbf{Maximum Speed (MaxS)}, when associated with a collision, denotes the maximum speed of the involved vehicles before the crash starts until the full stop (e.g., between time points $t_1$ and $t_4$ in Fig.\ref{fig:safety metircs} 
MaxS is a simple but effective measure directly related to the severity of the collision.

\textbf{Differential Speed ($\Delta s$)} is defined as the relative speed between the involved vehicles. It occurs at $t_2$ in Fig.\ref{fig:safety metircs}.
The system needs to calculate the speed of the involved vehicles (using methods like DL-based object tracking with or without explicitly extracting the motion trajectories) to determine MaxS and $Delta s$.
\begin{figure}[]
\begin{center}
\centerline{\includegraphics[width=0.9\columnwidth]{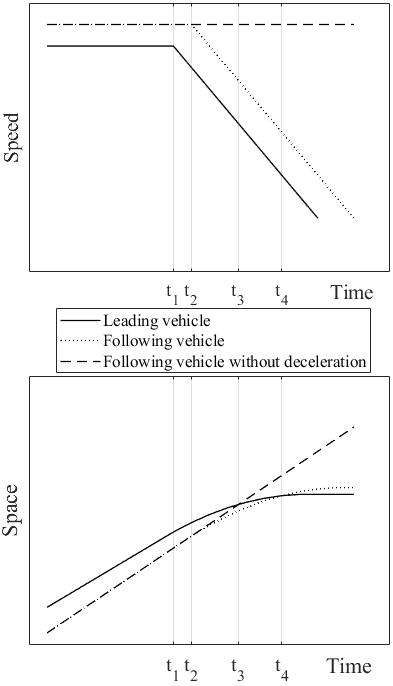}}
\caption{The space-speed pair diagram of a rear-end collision. $t_1$ is the time epoch the leading vehicle begins to decelerate (encroachment begin). $t_2$ is the epoch the following vehicle begins to decelerate. $t_1$-$t_2$ is the reaction time of the driver of the following vehicle. $t_3$ is the projected arrival moment of the following vehicle (under no deceleration). $t_4$ is the actual collision moment.}
\label{fig:safety metircs}
\end{center}
\end{figure}

\textbf{Time to Collision (TTC)} is a commonly used surrogate measure to define the time of an upcoming rear-end collision between two vehicles, if they continue their current speed ($t_2$-$t_3$  in Fig. \ref{fig:safety metircs}). The system needs to observe the relative position and calculate the relative velocity of the two involved vehicles. 
TTC is computed as:

\begin{align}
\label{eq:TTC}
    TTC =\frac{X_L-X_F}{v_F-v_L},
\end{align}

where $X_i$ denotes the position, $v_i$ denotes the velocity, and the indexes $L$ and $F$, respectively, denote the leading and following vehicles. The collision occurs only if $v_F\geq v_L$. Usually, we have $v_L \geq 0$ assuming that the cars move in the same direction. $v_L \leq 0$ represents a vehicle approaching from the front. The severity of an encounter can be determined by the minimum TTC (TTCmin). 
Although very useful in investigating crashes, this metric has some limitations. First, TTC assumes that involved vehicles are moving at constant speeds, which ignores the potential dangers caused by acceleration or deceleration. Secondly, it assumes that the cars drive in the same direction and not appropriate for side crashes.

\textbf{Post-Encroachment Time (PET)} is defined as the time span between the encroached vehicle leaving and the other vehicle with the right-of-way arriving at the conflict point. The system needs to observe the relative time of the involved vehicles and estimate the conflict point. 
PET in Fig.\ref{fig:PET} is computed as:

\begin{align}
\label{eq:PET}
    PET =t_b-t_a,  
\end{align}
where $t_a$ and $t_b$ represent the arrival time of the two involved vehicles. 
PET can be easily captured and computed by the system. However, it is suitable for intersection conflicts with transversal trajectories, but it does not fully reflect the severity of the crash. 

\begin{figure}[]
\begin{center}
\centerline{\includegraphics[width=1\columnwidth]{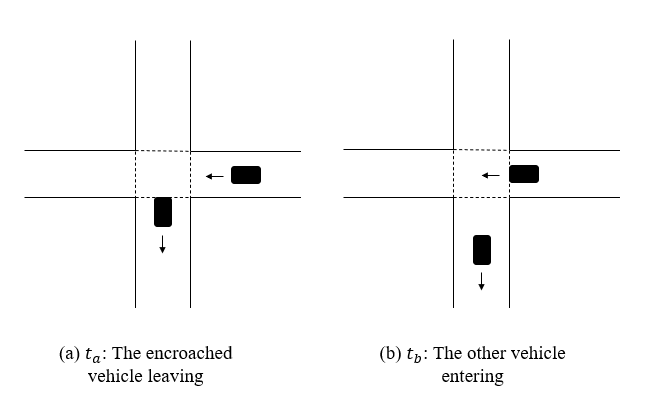}}
\caption{The schematic diagram of PET.}
\label{fig:PET}
\end{center}
\end{figure}

\textbf{Initial Deceleration Rate (Initial DR)} quantifies the avoidance behavior taken by a vehicle to avoid a collision. More specifically, Initial DR (the second derivative of Curve B at time point $t_2$ in Fig.\ref{fig:safety metircs}) is defined as the deceleration rate $a_0$ at the beginning of the decelerating state. Deceleration rate is an appreciated variable to assess the potential severity of a conflict. Other variants of this metric family include Deceleration Rate to Avoid a Crash (DRAC), Crash Potential Index (CPI), and Criticality Index Function (CIF). 

\textbf{Deceleration Rate to Avoid a Crash (DRAC)} is defined as the minimum deceleration rate of the following vehicle to avoid a crash to the leading vehicle. 
To characterize this metric, the observer system should keep track of the relative positions and the relative velocities of the two vehicles. Note that DRAC fails to evaluate lateral movements and is applicable only to scenarios where both cars are on the same lane. Mathematically, DRAC is defined as 

\begin{align}
\label{eq:DRAC}
    DRAC =\frac{(v_F-v_L)^2}{X_L-X_F}. 
\end{align}

\textbf{Crash Potential Index (CPI)} is defined as the probability of DRAC exceeding the Maximum Available Deceleration Rate (MADR) at any moment. 
MADR depends on the type of the vehicle as well as the environmental conditions. The system needs to calculate MADR of the target vehicles and instantaneous track time-variant deceleration to obtain the 
crash likelihood. CPI is defined as 

\begin{align}
    CPI_i=\frac{\sum_{t=t_i}^{tf_i}P(DRAC_{i,t})\geq MADR ~\Delta t\cdot b}{T_i},
\end{align}
where $t_i$ and $tf_i$ are the initial and final time intervals of vehicle $i$, respectively, $T_i$ denotes the total travel time of vehicle $i$, $\Delta t$ is the observation time step, and $b$ is a binary state variable with $b=1$ for active state and $b=0$ for inactive state. Compared with DRAC, CPI considers more factors such as the traffic and road conditions, but it is not yet suitable for assessing lateral movements.

\textbf{Proportion of Stopping Distance (PSD)} is defined as the ratio of the Remaining Distance (RD) from the current position to the potential collision point to the acceptable Minimum Stopping Distance (MSD).
It can be stated as

\begin{align}
\label{eq:PSD}
    PSD =\frac{RD}{MSD}.
\end{align}

If MSD is considered under maximum deceleration, then PSD becomes:

\begin{align}
\label{eq:MSD}
    PSD =\frac{v^2}{2 MADR}.
\end{align}

PSD is an easily observable indicator and is one of the few metrics that can be used for a conflict involving a single vehicle (e.g., crashing into a stationary obstacle like a tree). PSD is defined for evasive actions and usable only for specific safety problems. \ar{If PSD is large, one can say the situation is safe, but a small PSD doesn't necessarily indicate a high crash risk \cite{mohammadian2021integrating}.}

\textbf{Unsafe Density (UD)} is defined as the severity of the potential crash when the leading vehicle is within the achievable maximum DR. This metric is introduced in \cite{torday2003indicator}. When multiple cars involved, this metric considers all cars in a link, while the similar metric of UnSafety (UNS) considers only two of the cars from the link to calculate the $UNS$ of car $v$ at time step $s$, denoted by $UNS_{v,s}$. 

According to \cite{torday2003indicator}, the severity of a read-end crash is proportional to $\Delta v$, and $v_F$. 
More specifically, UNS is defined as:

\begin{align}
    UNS = \Delta v\cdot v_F\cdot R_d,
\end{align}
where $R_d$ denotes the ratio between the deceleration of the leading vehicle and its maximum deceleration capacity, namely

\begin{equation}
    R_d=
\begin{cases}
b/b_{max}& \text{b<0}\\
0& \text{else}
\end{cases}.
\end{equation}
Then, UD can be written as:

\begin{align}
    UD=\frac{\sum_{S=1}^{S_t}\sum_{V=1}^{V_t} UNS_{V,S}\cdot d}{TL},
\end{align}
where $S_t$ denotes the number of simulation steps within the aggregation period, $V_t$ denotes the number of vehicles in the link, $d$ denotes the time span between the two simulation steps, $T$ denotes the aggregation period, and $L$ denotes the section length for which the metrics are evaluated. 

UD provides more accurate information than typical micro-simulation outputs and allows comparison between the simulation scenarios or links. In a transportation network, each intersection is considered as a node, and the traffic flow between the nodes is represented by a link. 
The two limitations of the UD parameter include (i) the difficulty of quantifying its constituent parameters and (ii) its applicability only to identical trajectories, namely when two cars move in the same direction, and the type of the potential crash is rear-end collision.

\subsection{Operational safety metric for ADS proposed by the IAM} 
Here, we review part of the recently proposed metrics by the IAM \cite{wishart2020driving}. These metrics are developed by a team whose leader is a co-author of this paper. 

\textbf{Minimum Safe Envelope (MSE)} is defined to indicate the minimum longitudinal and lateral distances that the subject vehicle should maintain from other safety-relevant entity (often is another vehicle) for safety purposes. When the subject vehicle (subscript 1) is following behind another entity (subscript 2) and both of them are moving in the same direction. The longitudinal boundary can be defined as:

\begin{align}\label{eq:d_long_same}
    d_{min}^{long,same} &= v_1^{long}\rho_1+\frac{1}{2}a_{1,max,acc}^{long}\rho_1^2\\\nonumber
    &+ \frac{(v_1^{long}+a_{1,max,acc}^{long}\rho_1)^2}{2a_{2,min,dec}^{long}}\\\nonumber
    &-\frac{(v_2^{long})^2}{2a_{2,max,dec}^{long}},
\end{align}
where $v_i^{long/lat}$ denotes the current longitudinal/lateral velocity of  vehicle $i$, $a_{i,max/min,acc/dec}^{long/lat}$ denotes the longitudinal/lateral maximum/minimum acceleration/deceleration of vehicle $i$, and $\rho_i$ denotes the response time of vehicle $i$. 
When two involved vehicles are moving in opposite directions towards each other, the longitudinal boundary can be defined as: 

\begin{align}\label{eq:d_long_opp}
    d_{min}^{long,opp} =& v_1^{long}\rho_1+\frac{1}{2}a_{1,max,acc}^{long}\rho_1^2\\ \nonumber
    +& \frac{(v_1^{long}+a_{1,max,acc}^{long}\rho_1)^2}{2a_{1,min,dec}^{long}}\\ \nonumber
    +& \vert v_2^{long}\vert\rho_2+\frac{1}{2}a_{2,max,acc}^{long}\rho_2^2\\\nonumber
    +& \frac{(\vert v_2^{long}\vert+a_{2,max,acc}^{long}\rho_2)^2}{2a_{2,min,dec}^{long}}
\end{align}
The lateral boundary can be defined as:

\begin{align}\label{eq:d_lat}
    d_{min}^{lat} = \mu+ & v_1^{lat}\rho_1+\frac{1}{2}a_{1,max,acc}^{lat}\rho_1^2\\\nonumber
    +& \frac{(v_1^{lat}+a_{1,max,acc}^{lat}\rho_1)^2}{2a_{1,min,dec}^{lat}} \\\nonumber
    - & [v_2^{lat}\rho_2-\frac{1}{2}a^{lat}_{2,max,acc}\rho_2^2\\\nonumber
    -& \frac{(v_2^{lat}-\rho_2 a_{2,max,acc}^{lat})^2}{2a_{2,min,dec}^{lat}} ],
\end{align}
where $\mu$ is the lateral fluctuation margin. 
Noting that if the calculated result of $d_{min}^{long,same}$ or $d_{min}^{lat}$ is negative, it should be rounded up to 0. 
While the two boundaries are violated by the subject vehicle, MSDV is active ($MSDV=1$), meaning that 
an avoidable accident may occur. This metric characterizes the quantified risky distances as the basis of many assessment methods.

\textbf{Proper Response (PR)} is defined to indicate a proper action taken by the subject vehicle to recover itself when the MSE's safety boundaries ($d_{min}^{long},d_{min}^{lat}$ and the safe range of $a^{lat}$,$a^{long}$) are violated. It adds more information beyond MSE to evaluate the behavior of the subject vehicle.  

\textbf{Minimum Safe Distance Factor (MSDF)} is defined as the ratio between the current distance[s] to the calculated safe boundaries from the surrounding entity.
It can be found as:

\begin{align}
    &MSDF^{lat}& = \frac{d^{lat}}{d^{lat}_{min}}, &  MSDF^{long}& = \frac{d^{long}}{d^{long}_{min}},
\end{align}
where $d^{lat}$ and $d^{long}$ denote the measured distances. $MSDF\geq 1$ indicates the degree of defensive driving style. Note that a higher $MSDF$ may not mean a higher safety, necessarily.

\textbf{Minimum Safe Distance Calculation Error (MSDCE)} is defined as the difference between the calculated results by ADS from the ground truth. It can be formulated as:

\begin{align}
    MSDCE = \sqrt{\frac{\vert d^{long}_{gt,min} - d^{long}_{min}\vert }{d^{long}_{gt,min}}+\frac{\vert d^{lat}_{gt,min} - d^{lat}_{min}\vert }{d^{lat}_{gt,min}}}
\end{align}

This metric is derived from the MSDV, which indicates the ADS ability to determine the safety distances.

\textbf{Collision Incident (CI)} is defined to indicate the subject vehicle is involved in a collision determined by the reasonably related data. CI is active when $d^{lat}$ and $d^{long}$ are equal to 0. Also, the severity of the collision can be defined by KABCO scale\cite{herbel2010highway} with K(Fatal Injury), A(Incapacitating Injury), B (Non-Incapacitating Injury), C(Possible Injury), O(No Injury). 

\textbf{Traffic Law Violation (TLV)} is defined to indicate the subject vehicle that violates a traffic law which would result in an infraction or citation. Violating these laws sets TLV active. This metric emphasizes that an ADS-vehicle must follow existing laws, and an active TLV may lead to severe safety issues. It should be noted that exceptions to a TLV are made, for example, when road closures require temporary violations of driving exclusively inside a traffic lane.

\textbf{Achieved Behavioral Competency (ABC)} is defined to indicate the subject vehicle can execute a specific behavior correctly. This metric indicates the safety of ADS and is included in the preliminary list of ADS vehicles' metrics.

\textbf{ADS Active (ADSA)} is defined to indicate that the ADS is active when executing behaviors. This metric represents the safety of ADS and is included in the preliminary list of ADS vehicles' metrics. The metric has some dispute in California\cite{hawkins2019california}.

\textbf{Human Traffic Control Detection Error Rate (HTCDER)} is defined as the capability to detect instructions from a Human Traffic Control (HTC) actor correctly. It is calculated as:

\begin{align}
    HTCDER = \frac{GTI-CDI}{CDI},
\end{align}
where $GTI$ is the number of ground truth instructions, and $CDI$ is the number of correctly detected instructions. This metric is one of the HTC measurements to evaluate the safety of the subject vehicle. Note that the manual instructions by officers should still be considered in this metric.

\textbf{Human Traffic Control Violation Rate (HTCVR)} is defined as the capability of a subject vehicle to follow the received instructions successfully. It is formulated as:

\begin{align}
    HTCVR = \frac{CDI-CCI}{CDI},
\end{align}
where $CCI$ is the number of correctly compiled instructions. 

\textbf{Aggressive Driving (AD)} is defined to indicate the maneuvers (longitudinal/lateral accelerations) of a subject vehicle exceeding specified thresholds. When exceeding ($a\geq a_T$), the metric is set active. The thresholds vary by jurisdiction and culture. This metric involves only the subject vehicle and implies the inherent and potential risks of natural driving behaviors. When evaluating the safety of ADS, this metric should be included.


\fi





\end{document}